\PassOptionsToPackage{numbers, compress, sort}{natbib}
\documentclass{article}
\usepackage[final]{neurips_2020}
\usepackage[utf8]{inputenc}
\usepackage[T1]{fontenc}
\usepackage{url}
\usepackage{booktabs}
\usepackage{amsfonts}
\usepackage{nicefrac}
\usepackage{microtype}
\usepackage{amsmath,amssymb}
\usepackage{graphicx}
\usepackage[backgroundcolor=white,linecolor=red,bordercolor=none,textwidth=3em,textsize=footnotesize]{todonotes}
\usepackage[colorlinks=true]{hyperref}
\usepackage{amsmath,amssymb,xspace,graphicx,bm}
\usepackage[capitalise]{cleveref}
\usepackage{multirow}
\usepackage{relsize}
\usepackage{wrapfig}
\usepackage{subcaption}
\usepackage{appendix}
\usepackage{titles}[appendix]
\crefname{section}{Sec.}{Sec.}
\crefname{equation}{Eq.}{Eq.}
\crefname{table}{Tab.}{Tab}
\crefname{appendix}{Appendix}{Appendix}

\newcommand{\ie}{\textit{i}.\textit{e}.~}
\newcommand{\eg}{\textit{e}.\textit{g}.~}

\DeclareMathOperator*{\argmin}{arg\,min}

\makeatletter
\renewcommand{\paragraph}{%
  \@startsection{paragraph}{4}%
  {\z@}{0em}{-1em}%
  {\normalfont\normalsize\bfseries}%
}
\makeatother

\title{Quantifying Learnability and Describability of Visual Concepts Emerging in Representation Learning}

\author{%
  Iro Laina
  \qquad
  Ruth C. Fong
  \qquad
  Andrea Vedaldi
  \vspace{1em} \\
  Visual Geometry Group \\
  University of Oxford \\ 
  \texttt{\{iro, ruthfong, vedaldi\}@robots.ox.ac.uk}
}

\begin{document}

\maketitle
\begin{abstract}
The increasing impact of black box models, and particularly of unsupervised ones, comes with an increasing interest in tools to understand and interpret them.
In this paper, we consider in particular how to characterise visual groupings discovered automatically by deep neural networks, starting with state-of-the-art clustering methods.
In some cases, clusters readily correspond to an existing labelled dataset.
However, often they do not, yet they still maintain an ``intuitive interpretability''.
We introduce two concepts, visual learnability and describability, that can be used to \emph{quantify} the interpretability of arbitrary image groupings, including unsupervised ones.
The idea is to measure (1) how well humans can learn to reproduce a grouping by measuring their ability to generalise from a small set of visual examples (learnability) and (2) whether the set of visual examples can be replaced by a succinct, textual description (describability).
By assessing human annotators as classifiers, we remove the subjective quality of existing evaluation metrics.
For better scalability, we finally propose a class-level captioning system to generate descriptions for visual groupings automatically and compare it to human annotators using the describability metric.
\end{abstract}

\section{Introduction}\label{s:intro}

Recent advances in unsupervised and self-supervised learning have shown that it is possible to learn data representations that are competitive with, and sometimes even superior to, the ones obtained via supervised learning~\cite{misra2020self,he2020momentum}.
However, this does not make unsupervised learning a solved problem;
unsupervised representations often need to be combined with labelled datasets before they can perform useful data analysis tasks, such as image classification.
Such labels induce the semantic categories necessary to provide an \emph{interpretation} of data that makes sense to a human.
Thus, it remains unclear whether unsupervised representations develop a human-like understanding of complex data in their own right.

In this paper, we consider the problem of assessing to what extent abstract, human-interpretable concepts can be discovered by unsupervised learning techniques.
While this problem has been looked at before, we wish to cast it in a more principled and general manner than previously done.
We start from a simple definition of a \emph{class} as a subset $\mathcal{X}_c \subset \mathcal{X}$ of patterns (\eg images).
While our method is agnostic to the class generation mechanism, we are particularly interested in classes that are obtained from an unsupervised learning algorithm.
We then wish to answer three questions:
(1) whether a given class is interpretable and coherent, meaning that it can be indeed understood by humans,
(2) if so, whether it is also describable, \ie it is possible to distill the concept{}(s) that the class represents into a compact sentence in natural language, and
(3) if such a summary description can be produced automatically by an algorithm.

The first problem has already been explored in the literature (\eg \cite{zhou2014object}), usually using human judgment to assess class interpretability.
In short, human annotators are shown example patterns from the class and they are asked to name or describe it.
Unfortunately, such a metric is rather subjective, even after averaging responses by several annotators.
In our work, we aim to minimize subjectivity in evaluating class interpretability.
While still involving humans in the assessment, we cast the problem as the one of \emph{learning} the class from a number of provided examples.
Rather than asking annotators to identify the class, we test their ability to discriminate further examples of the class from non-class examples.
The accuracy from this classification task can then be used as an objective measure of human learnability of the class, which we call \emph{semantic coherence}.
As we show later, self-discovered classes are often found to be semantically coherent according to this criterion, even when their consistency with respect to an existing label set\,---\,such as ImageNet~\cite{deng2009imagenet}\,---\,is low.

Note that our semantic coherence metric does not require naming or otherwise distilling the concept{}(s) captured by a class.
Thus, we also look at the problem of \emph{describing} the class using natural language.
We start by manually collecting names or short descriptions for a number of self-labelled classes.
Then, we modify the previous experiment to test whether annotators can correctly recognise examples of the class from negative ones based on the provided description.
In this manner, we can \emph{quantify} the quality of the description, which is directly related to how easily describable the underlying class is.

Finally, we ask whether the process of textual distillation can be automated; this problem is related to image captioning, but with some important differences.
First, the description has a well-defined goal: to teach humans about the class, which is measured by their ability to classify patterns based on the description.
This approach also allows a direct and quantitative comparison between manual and automatic descriptions.
Second, the text must summarise an entire class, \ie a collection of several patterns, rather than a single image.
To this end, we investigate ways of converting existing, single-image captioning systems to class-level captioning.
Lastly, we propose an automated metric to validate descriptions and compare such systems before using the human-based metric.

\begin{figure}[t]
    \centering
    \includegraphics[trim=0.8cm 7.35cm 0.8cm 0cm, clip, width=1.0\linewidth]{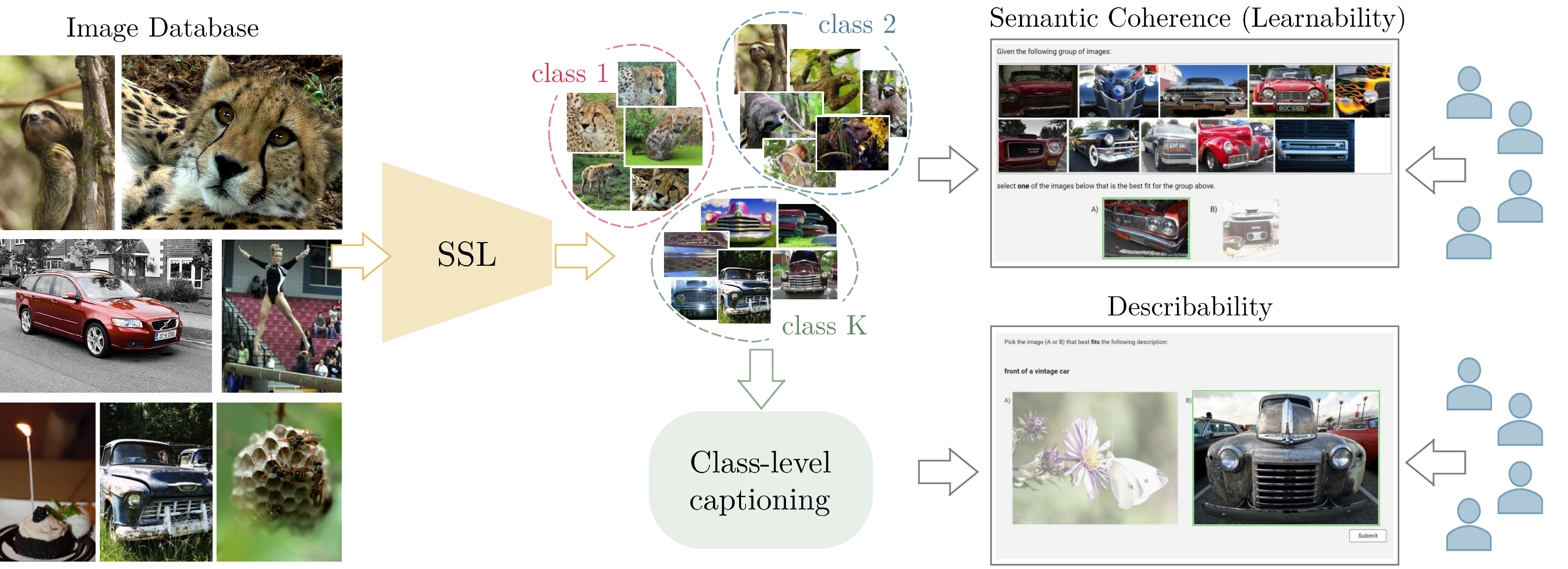}
    \caption{\textbf{Our framework}. We evaluate classes obtained by self-supervised learning algorithms (SSL) using human judgements. We formulate our evaluation as two forced-choice tasks and measure (a) the learnability of a class by humans and (b) the describability, \ie the ability to distill the gist of the class into a description in natural language (class-level captioning).}
    \label{fig:framework}
\end{figure}

\section{Related work}\label{s:related}

\paragraph{Unsupervised representation learning and self-labelling.}
In this work, we primarily study representations learned 
from unlabelled data.
A number of methods use domain-specific, pretext tasks to learn rich features~\cite{pathak2016context,doersch15unsupervised, noroozi16unsupervised,jing2020self}.
Contrastive learning is another promising direction that has recently lead to several state-of-the-art methods~\cite{wu2018unsupervised,bachman2019learning,he2020momentum,chen2020simple,oord2018representation,srinivas2020curl,tian2019contrastive,henaff2019data,hjelm2018learning}, closing the gap to supervised pretraining.
This paradigm encourages similar examples to be close together in feature space and non-similar ones to be far apart.
A third research direction combines representation learning and clustering, \ie jointly learning data representations and labels, in a variety of ways~\cite{ji2019invariant,hu17learning,yang2016joint,caron2018deep,caron2019unsupervised,yang2017towards,asano20self-labelling,bojanowski2017unsupervised,noroozi2018boosting,yan2020cluster,caron2020unsupervised}.
The representational quality of self-supervised methods is most commonly evaluated by performance on downstream tasks (\eg ImageNet~\cite{deng2009imagenet}), but
relatively little work has been done on characterizing the kinds of visual concepts learned by them.

\paragraph{Model interpretability.}
A large body of work has focused on understanding the feature representations of CNNs; these typically focus on supervised networks.
One approach is to \emph{visualise} feature space dimensions (\eg a single filter or a linear combination of filters).
This can be done using real examples, \ie by showing the top image patches that most activate along a given direction~\cite{zeiler14visualizing,zhou2014object},
or by using generated examples
(a.k.a. activation maximization or feature visualization)~\cite{simonyan14deep,mahendran16visualizing,olah17feature,nguyen16synthesizing,nguyen2017plug,ulyanov18deep}.
Another approach is to \emph{label} dimensions;
this can be done automatically by correlating activations with an annotated dataset of semantic concepts~\cite{bau17network,fong18net2vec,kim17interpretability,zhou2018interpretable}. In particular,~\cite{bau17network,fong18net2vec} use this paradigm to compare supervised representations with self-supervised ones.
This can also be done by asking human annotators to label or name examples
via crowd-sourcing platforms like Amazon Mechanical Turk~\cite{zhou2014object,gonzalez2018semantic}.
Annotators can be used to compare the interpretability of different visualizations by asking them to choose the visualization they prefer~\cite{zhou2018interpretable}.
Our work is most similar to~\cite{zhou2014object,gonzalez2018semantic}, yet we differ in that we replace the subjective nature of asking for free-form labels with an objective task that the annotators are asked to perform.

\paragraph{Learning visual concepts.}
Our approach to understanding the visual concepts learned by representation learning algorithms is to test whether they are ``learnable'' by humans.
The work by \citet{jia2013visual} is similar in that they ask humans whether query images belong to a reference set of a visual concept.
The human judgements then serve as the ground-truth signal to a learning system.
In contrast, we use human judgements to assess the learnability of \emph{machine}-discovered concepts.
Relatedly, \cite{zhou2019humans} measure the relatedness of human judgements and machine predictions in the context of adversarial images.

\paragraph{Related topics in cognitive science.}
Several works in cognitive science have also studied the learnability of concepts by humans, starting from word learning: \citet{xu2007word} let human subjects observe visual examples of a novel word and estimate the probability of a new example being identified as the same word/concept.
This problem is closely related to the notion of representativeness~\cite{tenenbaum2001rational}, which is addressed in \cite{abbott2011testing,griffiths2016exploring} using natural image datasets.
Work on representativeness, \ie estimating how well an image represents a set of images, aims to model human beliefs with Bayesian frameworks.
In the context of behavioral studies, \cite{hsu2012identifying} further attempt to couple Monte Carlo Markov Chains (MCMC) with people, through a series of forced choice tasks of selecting an image that best fits a description.
The oddball paradigm~\cite{bravo1992role}, which asks subjects to pick the ``odd'' example out of a set of examples, is also related to our work.

\paragraph{Image captioning.}
Finally, we study the degree to which self-discovered concepts can be \emph{described} automatically.
This is similar to image captioning, 
except that the goal is to describe categories collectively, which requires reasoning about intra- and inter-class variation.
While some methods focus on discriminability~\cite{chen2018groupcap,luo2018discriminability,vedantam2017context, dai2017contrastive,liu2018show}, this is only done on image level, with the goal of generating more diverse captions.
Others aim to explicitly describe the differences between pairs of examples (\eg video frames)~\cite{park2019robust,JhamtaniB18,forbes2019neural,gilton2020detection}.
Most similar to our work is~\cite{li2020context}, which generates a short description for a target \emph{group} of images in contrast to a reference group.
Also related is the concept of visual denotation in \cite{young2014image}, \ie the set of images that depict a single linguistic expression.

\section{Coherence measures}\label{s:consistency}
In this section, we introduce our measure of semantic coherence and describability for visual classes.
Let $\mathcal{X}$ be a space of patterns, \eg a collection of natural images, and  $\mathcal{X}_c \subset \mathcal{X}$ be a given class.
We construct $\mathcal{X}_c$ as follows: Given a learned, binary function $\phi_c(x) \in \{0,1\}$, we define $\mathcal{X}_c = \{x \in \mathcal{X} : \phi_c(x)=1\}$ as the space of input images that ``activate'' $\phi_c$.
For deep clustering methods~\cite{caron2018deep,asano20self-labelling}, which assign images to clusters, $\phi_c$ is an indicator of assignment to cluster $c$.
For unsupervised representation learning methods~\cite{he2020momentum,chen2020simple},
$\phi_c$ can be constructed by clustering the learned representations.
For an arbitrary neural network, one can define $\phi_c$ by thresholding activations of filter $c$ in a given layer, as done in~\cite{zhou2014object,bau17network}.

\paragraph{Class learnability and semantic coherence.}

Previous work in deep clustering~\cite{caron2018deep} suggests that the learned clusters
are not exclusively object categories but often exhibit more abstract concepts, patterns, or artistic effects, 
which cannot be captured by comparing self-supervised representations to annotated datasets (\eg ImageNet).
Consequently, we wish to assess semantic groupings independent from any a-priori data labelling.

We formulate this as testing whether a class $\mathcal{X}_c$ is \emph{semantically coherent} by testing if the class can be easily \emph{learned} by humans.
For this purpose, we do not require the class to be easily describable via natural language; instead, we  show the annotators examples from the class and ask them to classify further images as belonging to the same class or not, measuring the classification error.

Formally, given the class $\mathcal{X}_c$, we conduct a number of human-based tests in order to assess its learnability or semantic coherence.
Each test $T$ is a tuple $(\hat{\mathcal{X}}_c,x_0,x_1,z,h \mid \mathcal{X},\mathcal{X}_c)$ where $\hat{\mathcal{X}}_c \subset\mathcal{X}_c$ is a random subset of images from class $c$ with a fixed cardinality (\ie $|\hat{\mathcal{X}}_c|=M$).
Then, $x_0\in\mathcal{X}_c-\hat{\mathcal{X}}_c$ is a random sample of class $c$ not in the representative set $\hat{\mathcal{X}}_c$,
and $x_1\in\mathcal{X}-\mathcal{X}_c$ is a random sample that does not belong to class $c$ (\ie background).
Finally, $z \in \{0,1\}$ is a sample from a uniform Bernoulli distribution and $h$ is a human annotator, selected at random from a pool.

The human annotator $h$ is presented with the sample $(\hat{\mathcal{X}}_c,x_z,x_{1-z})$ via a user interface and her/his goal is to predict the value of $z$:
$
  \hat z = h(\hat{\mathcal{X}}_c,\, x_z,\, x_{1-z}).
$
Intuitively, the annotator is shown a number of images $\hat{\mathcal{X}}_c$ from the class as reference and two query images $x_0$ and $x_1$, one which belongs to the class and one which does not, in randomized order.
The annotator's goal is to identify which of the two images belongs to the class.
We then define the \emph{coherence} of the class as
\begin{equation}\label{e:coherence}
  C(\mathcal{X}_c) = \mathbb{E}_T[z = h(\hat{\mathcal{X}}_c,\, x_z,\, x_{1-z})],
\end{equation}
that is the average accuracy of annotators in solving such tasks correctly.
We thus evaluate the average ability of annotators to learn the class from the provided examples.

Finally, we consider different functions for sampling negatives which correspond to exploring different aspects of the learned classes. 
Specifically, we consider two different functions. 
The first and simplest choice is to sample negatives uniformly at random from $\mathcal{X} - \mathcal{X}_c$. 
This approach explores the overall learnability of a class. 
The second is to provide the annotator with a binary choice between a positive and a \emph{hard negative} query. 
The hard negative image can be sampled from another class $\mathcal{X}_{c^*}$, where 
\begin{equation}
c^* = \argmin_{c' \ne c} 
\left| 
\alpha(\mathcal{X}_{c}) - \alpha(\mathcal{X}_{c'})
\right| 
\quad\mathrm{and}\quad
\alpha(\mathcal{X}) = \frac{1}{|\mathcal{X}|} \sum_{x \in \mathcal{X}} f(x) ~.
\end{equation}
In the above equation, $f(x)$ is a function producing a feature representation of image $x$.
In other words, we sample hard negatives only from the class that is ``most similar'' to the target one based on class centroids in the feature space induced by $f$. 
Intuitively, this approach tests whether there are sufficient fine-grained differences between classes to be learnable by humans.

The advantage of \cref{e:coherence}, when compared to alternatives presented in prior work~\cite{zhou2014object,gonzalez2018semantic}, is that it does not require the class to be easily describable in words.
It also provides a simple, testable and robust manner to assess the visual consistency of any image collection, including self-supervised classes. 
Note that there are more variants of the problem above.
First, the problem presented to the annotator can be modified in various ways, \eg presenting more queries.
In particular, it is possible to also show explicitly more than one ``negative'' example (in our case, they are shown exactly one, $x_1$, v.s. $M+1$ positive examples).
Second, the difficulty of the problem can be modulated by the ambient space, and hence background images, differently.
If all images in $\mathcal{X}$ are similar (\eg only dogs), then separating a particular class $\mathcal{X}_c$ (\eg a dog breed) is significantly harder than if $\mathcal{X}$ is more generic (\eg random Internet images).
We leave the exploration of these extensions to future work.

\paragraph{Class describability and description quality.}

We are also interested in assessing whether self-supervised classes capture concepts that can be compressed into a natural language phrase that describes the gist of the class.
Such concepts might be represented by higher-level semantics, such as object categories or scenes (\eg \emph{puppy lying on grass}) or actions (\eg \emph{racing}), but they can also refer to mid-level characteristics (\eg \emph{striped texture}).

We extend the idea of semantic coherence to also include a free-form, language-based description for the class.
The assessment task is the same as above, but instead of seeing a reference set of  examples from a given class, annotators are shown a short description of the class.
Prior work characterizes the describability of a class by asking annotators whether they believe a given class is semantic; this is a subjective assessment.
Instead, we measure the effectiveness of a description in characterizing a class by its ability to convey useful information, \ie to ``teach'' a human annotator about a class; this is an objective assessment of the utility of a description.

Formally, the protocol mentioned above is modified by replacing $\mathcal{X}_c$ with a description $D_c$ of the class $\mathcal{X}_c$ in natural language.
We evaluate the \emph{describability} of the class as:
\begin{equation}\label{e:description-coherence}
  C(\mathcal{X}_c,\,D_c) = \mathbb{E}[z = h(D_c,\, x_z,\, x_{1-z})].
\end{equation}
\cref{e:description-coherence} captures the effect of two factors.
The first one is the semantic coherence of the underlying class $\mathcal{X}_c$.
When class samples exhibit low coherence, the class can neither be understood by annotators nor described precisely and compactly.
The second factor is the quality of the description $D_c$ itself.
Namely, given a fixed class $\mathcal{X}_c$, we can use~\cref{e:description-coherence} to assess different descriptions $D_c$ based on their efficacy in covering the information required to characterize the class.
As a baseline in this experiment, we consider human-generated descriptions, which helps decouple the two factors.

\section{Automatic class-level captioning}\label{s:method}

Given the above protocol for assessing class describability with human subjects, we next consider the problem of generating descriptions for arbitrary image collections automatically.
We hereby refer to this problem as \emph{class}-level captioning, emphasising the difference from \emph{image}-level captioning.
In particular, the goal in class-level captioning is to accurately describe not just a single image, but the entire collection\,---\,or the most representative part of it\,---\,which requires distilling the commonalities of images that fall under the given collection.
Moreover, to encourage discriminativeness across various classes, such descriptions must be as specific as possible.
For example, the description ``organism'' may accurately describe a class; however, it likely can also be applied to more than one class, and is thus an inadequate description.

As there exist no training data for class-level captioning, our approach draws inspiration from unsupervised text summarisation techniques~\cite{mihalcea2004textrank,nenkova2012survey}.
As a first step, we use a pre-trained captioning model $g: \mathcal{X} \rightarrow \mathcal{S}$ to generate descriptions $s = g(x)$ for each image $x \in \mathcal{X}$ independently.
We then %
find the most representative description for each class from 
$\mathcal{S}_c = \{g(x) \mid x \in \mathcal{X}_c\} \subset \mathcal{S}$: %
\begin{equation}
\label{eq:cap_opt}
    D_c = \argmin_{s \in S_c} \left(
    \frac{1}{|\mathcal{S}_c|}
    \mathlarger{\sum}_{s^+ \in \mathcal{S}_c \setminus \{s\}} \!\!\!d(s, s^+) \,-\,
    \frac{1}{|\mathcal{S} - \mathcal{S}_c|}
    \mathlarger{\sum}_{s^- \in \mathcal{S} \setminus \mathcal{S}_c} d(s, s^-)
    \right).
\end{equation}
In the above optimization, $d(\cdot,\cdot)$ is a distance metric between pairs of captions.
Intuitively, we choose a caption that is close to other captions for a given class while simultaneously being far away from captions of other classes. This is done by selecting the caption that maximizes the difference between the intra-class ($\mathcal{S}_c$) and inter-class ($\mathcal{S}\setminus\mathcal{S}_c$) average caption distance.

We note that any metric suitable for evaluating language tasks can be used, such as ROUGE~\cite{lin2004rouge}, which is commonly used in text summarisation.
However, in order to better account for semantic similarities present in the captions, we define our distance function as
$
    d(s, s') = 1-\frac{\psi(s)^T \psi(s')}{\|\psi(s)\|~\|\psi(s')\|},
$
where $\psi: \mathcal{S} \rightarrow \mathbb{R}^n$ is a function mapping a sentence to an $n$-dimensional embedding space.
This allows for sentences that have common semantic properties to be represented by similar vectors.
Then, $d$ computes the cosine distance between two sentences in embedding space.
We can obtain $\psi(\cdot)$ from the captioning model itself, or, in the general case, we can use an off-the-shelf sentence encoder that captures semantic textual similarities~\cite{kiros2015skip,reimers2019sentence, cer2018universal}.

In contrast to image captioning, we do not evaluate our automatic descriptions directly against human-provided descriptions, due to known limitations of evaluation metrics for this task~\cite{cui2018learning, KilickayaEIE17}.
Instead, here we can evaluate both automatic and human-generated descriptions using our describability metric, which measures how effective a description is in teaching humans to classify images correctly.

\section{Experiments}\label{s:experiments}

Our experiments are organized as follows.
First, we examine the representations learned by two state-of-the-art approaches, namely SeLa~\cite{asano20self-labelling} and MoCo~\cite{he2020momentum}, and use our learnability metric (\cref{e:coherence}) to quantify the semantic coherence of their learned representations.
We then repeat theses experiments by providing human-annotated, class-level descriptions to measure the respective describability.
We further validate the approach against selected ImageNet categories, which are highly-semantic by construction and for which an obvious description (\ie the object class name) is readily available.
Finally, we evaluate the automatic class-level descriptions that we obtain from our method.

\subsection{Assessing unsupervised image clustering}

\paragraph{Collection of human judgments.}
To conduct experiments with human participants, we use Amazon Mechanical Turk (AMT).
For each method (SeLa, MoCo, ImageNet), we select a number of classes and create 20 Human Intelligence Tasks (HITs) for each class. 
Each HIT is answered by 3 participants.
In the following, we use data from the training set of ImageNet~\cite{deng2009imagenet}. 
For the semantic coherence experiments, each HIT consists of a reference set of 10 example images randomly sampled from the class and two query images.
The participants are asked to select the query image that is a better fit for the reference set.
For the describability HITs, we retain the same query images but replace the reference image set with the class description, which is either manual or automatic.
For the describability experiments, we restrict the number of HITs that a participant can answer to 1 per day \emph{per class} such that it is not possible to unintentionally learn the class from the answers.
Overall, 12k different HITs were answered by a total of 25,829 participants.

\paragraph{Evaluation metrics.}
The coherence of a class $\mathcal{X}_c$, as defined by \cref{e:coherence}, is the probability that annotators can identify the correct class out of a binary choice.
In order to provide a higher-level analysis of our results, we report $C(\mathcal{X}_c)$ averaged over certain groups of classes, described below. 
In addition, we compute the 95\% confidence intervals (CI) using the Clopper-Pearson method~\cite{clopper1934use}. 
We also report %
the inter-rater reliability (IRR) using Krippendorff's $\alpha$-coefficient~\cite{krippendorff2011computing}, where
$\alpha = 1$ indicates perfect agreement, while $\alpha = 0$ indicates disagreement. %
Thus, low inter-rater agreement suggests that, in the forced-choice test given to the participants, the correct answer is not obvious.

\paragraph{Semantic coherence on ImageNet classes.}
In order to validate our methodology, we first conduct experiments on ImageNet~\cite{deng2009imagenet}, for which manually labelled categories exist, yielding $\mathcal{X}^\text{IN}_c$, $c\in \{1,\dots,K\}$ with $K=1,\!000$.
As a sanity check, we report the average coherence over 20 selected ImageNet classes%
in the last row of~\cref{t:sela-coherence}.
ImageNet labels are by definition highly semantic and consistent. Thus, as expected, the agreement of human participants with the ground truth labels is very high, reaching average semantic coherence of $99.0\%$.

\begin{table}[t]
\centering 
\caption{Human assessment of semantic coherence for self-supervised methods \cite{asano20self-labelling, he2020momentum} and ImageNet. We evaluate the semantic coherence of self-supervised classes using random (R) and hard (H) negatives. Results grouped by purity range.}
\vspace{2pt}
\footnotesize
\begin{tabular}{l@{\hskip 10pt} c ccc@{\hskip 20pt} ccc}
\toprule
& & \multicolumn{3}{c@{\hskip 20pt}}{Semantic Coherence (R)} & \multicolumn{3}{@{}c}{Semantic Coherence (H)} \\
\cmidrule(lr@{19pt}){3-5}
\cmidrule(l@{-1pt}r){6-8}
Method & Purity range & Mean & 95\% CI & IRR & Mean & 95\% CI & IRR \\
\midrule
\multirow{7}{*}{SeLa~\cite{asano20self-labelling}} 
& $(0.3,\, 0.4 ]$ & 71.8 & [68.0, 75.4]  &  32.8 & 55.3 & [51.3, 59.4] & 7.4 \\
& $(0.4,\, 0.5 ]$ & 94.2 & [92.0, 95.9]  &  87.7 & 60.0 & [56.0, 63.9] & 14.9 \\
& $(0.5,\, 0.6 ]$ & 97.2 & [95.5, 98.3]  &  95.3 & 71.8 & [68.0, 75.4] & 31.4 \\
& $(0.6,\, 0.7 ]$ & 99.7 & [98.8, 100.0] &  99.5 & 63.2 & [59.2, 67.0] & 22.9 \\
& $(0.7,\, 0.8 ]$ & 98.0 & [96.5, 99.0]  &  94.9 & 65.3 & [61.4, 69.1] & 25.9 \\
& $(0.8,\, 0.9 ]$ & 99.8 & [99.1, 100.0] &  99.8 & 63.8 & [59.8, 67.7] & 22.3 \\
& $(0.9,\, 1.0 ]$ & 98.8 & [97.6, 99.5]  &  98.0 & 72.2 & [68.4, 75.7] & 36.6 \\
\midrule
\multirow{7}{*}{MoCo~\cite{he2020momentum}} 
& $(0.3,\, 0.4 ]$ &  89.8 & [87.1, 92.1]  & 79.6  & 56.8 &  [52.8, 60.8] & 9.2 \\
& $(0.4,\, 0.5 ]$ &  93.8 & [91.6, 95.6]  & 86.3  & 63.2 & [59.2, 67.0] & 14.4 \\
& $(0.5,\, 0.6]$  &  96.5 & [94.7, 97.8]  & 93.0  & 62.5 & [58.5, 66.4] & 25.8 \\
& $(0.6,\, 0.7 ]$ &  98.8 & [97.6, 99.5]  & 96.7  & 64.7 & [60.7, 68.5] & 24.4 \\
& $(0.7,\, 0.8 ]$ & 100.0 & [99.4, 100.0] & 100.0 & 63.5 & [59.5, 67.4] & 20.5 \\
& $(0.8,\, 0.9 ]$ &  99.5 & [98.5, 99.9]  & 98.0  & 64.3 & [60.4, 68.2] & 22.1 \\
& $(0.9,\, 1.0 ]$ &  99.2 & [98.1, 99.7]  & 98.9  & 74.8 & [71.2, 78.3] & 52.0 \\
\midrule
ImageNet & - & 99.0 & [98.3, 99.5] &  98.0 & - & - & -  \\ 
\bottomrule
\end{tabular}
\label{t:sela-coherence}
\end{table}

\begin{table}[!ht]
\centering 
\caption{Human assessment of describability for self-supervised methods \cite{asano20self-labelling, he2020momentum} and ImageNet. Class descriptions are either obtained manually (Human) or automatically (Auto). ImageNet descriptions are class names. Results grouped by purity range.}
\vspace{2pt}
\footnotesize
\begin{tabular}{l@{\hskip 10pt} c ccc@{\hskip 20pt} ccc}
\toprule
& & \multicolumn{3}{c@{\hskip 20pt}}{Describability (Human)} & \multicolumn{3}{@{}c}{Describability (Auto)} \\
\cmidrule(lr@{19pt}){3-5}
\cmidrule(l@{-1pt}r){6-8}
Method & Purity range & Mean & 95\% CI & IRR & Mean & 95\% CI & IRR \\
\midrule
\multirow{7}{*}{SeLa~\cite{asano20self-labelling}} 
& $(0.3,\, 0.4 ]$ & 84.2 & [81.0, 87.0] & 63.7 & 78.7 & [75.2, 81.9] & 44.5 \\
& $(0.4,\, 0.5 ]$ & 95.3 & [93.3, 96.9] & 90.1 & 93.3 & [91.0, 95.2] & 84.5 \\
& $(0.5,\, 0.6 ]$ & 94.3 & [92.2, 96.0] & 87.2 & 94.5 & [92.4, 96.2] & 88.0 \\
& $(0.6,\, 0.7 ]$ & 97.0 & [95.3, 98.2] & 96.0 & 96.0 & [94.1, 97.4] & 90.9 \\
& $(0.7,\, 0.8 ]$ & 95.8 & [93.9, 97.3] & 91.8 & 94.8 & [92.7, 96.5] & 88.7 \\
& $(0.8,\, 0.9 ]$ & 99.0 & [97.8, 99.6] & 98.2 & 98.2 & [96.7, 99.1] & 97.1 \\
& $(0.9,\, 1.0 ]$ & 98.8 & [97.6, 99.5] & 98.4 & 97.3 & [95.7, 98.5] & 94.6 \\
\midrule
\multirow{7}{*}{MoCo~\cite{he2020momentum}} 
& $(0.3,\, 0.4 ]$ & 78.3 & [74.8, 81.6] & 55.9 & 83.7 & [80.5, 86.5] & 60.4 \\
& $(0.4,\, 0.5 ]$ & 95.0 & [92.9, 96.6] & 87.9 & 90.7 & [88.1, 92.9] & 76.4 \\
& $(0.5,\, 0.6]$  & 96.5 & [94.7, 97.8] & 91.2 & 92.7 & [90.3, 94.6] & 84.7 \\
& $(0.6,\, 0.7 ]$ & 96.3 & [94.5, 97.7] & 91.9 & 96.8 & [95.1, 98.1] & 94.8 \\
& $(0.7,\, 0.8 ]$ & 98.0 & [96.5, 99.0] & 97.3 & 97.3 & [95.7, 98.5] & 95.0 \\
& $(0.8,\, 0.9 ]$ & 97.8 & [96.3, 98.8] & 96.6 & 98.7 & [97.4, 99.4] & 98.2 \\
& $(0.9,\, 1.0 ]$ & 98.5 & [97.2, 99.3] & 97.0 & 97.3 & [95.7, 98.5] & 95.9 \\
\midrule
ImageNet & - & 95.6 & [94.3, 96.7] & 91.1 &  - & - & -  \\ 
\bottomrule
\end{tabular}
\label{t:sela-describability}
\end{table}

\begin{figure}[!t]
\vspace{-1em}
    \centering
    \includegraphics[width=1.0\textwidth]{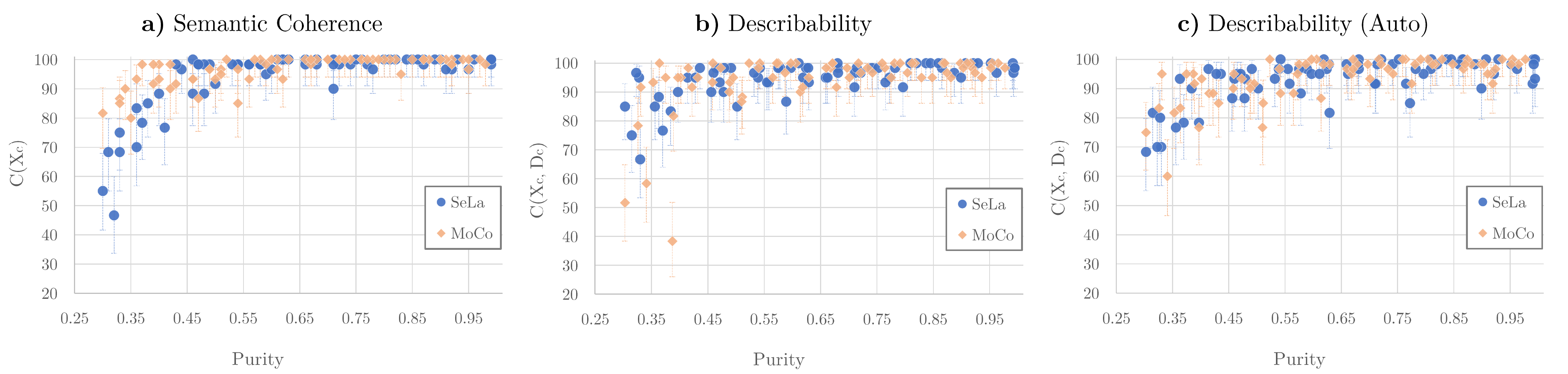}
    \vspace{-1.5em}
    \caption{\textbf{ImageNet purity vs. semantic coherence and describability}. Each point represents a SeLa~\cite{asano20self-labelling} or MoCo~\cite{he2020momentum} class evaluated on AMT. Semantic coherence is shown for randomly sampled negatives (for hard negatives, please see the Appendix). Error bars correspond to 95\% CI.}
    \label{fig:all_clusters}
\end{figure}

\paragraph{Semantic coherence on self-supervised classes.}
Next, we evaluate two state-of-the-art, self-supervised learning methods. 
SeLa~\cite{asano20self-labelling} simultaneously learns feature representations and predicts a clustering of the data by directly assigning images to clusters $\mathcal{X}_c^\text{SeLa}$ ($K=3000$).
We use the publicly available implementation
based on ResNet-50~\cite{he2016deep} and evaluate the clusters of the first out of 10 heads.
In contrast, MoCo~\cite{he2020momentum} does not produce its own labels, as it learns a continuous embedding space.
To obtain $\mathcal{X}_c^\text{MoCo}$, we apply $k$-means on top of MoCo-v1 feature vectors (obtained using the official implementation) %
and set $k=3000$ for a fair comparison with~\cite{asano20self-labelling}. 

We note that there is no a-priori relationship between self-supervised classes (\ie $\mathcal{X}_c^\text{MoCo}$ and $\mathcal{X}_c^\text{SeLa}$) and the human-annotated ones (\ie $\mathcal{X}^\text{IN}_c$).
We establish a relationship by computing the \emph{purity} of a class $\mathcal{X}_c$ as
$
    \Pi(\mathcal{X}_c)\!=\! 1 \!-\! \frac{H(l(\mathcal{X}_c))}{\log K},
$
where $l : \mathcal{X} \!\rightarrow\! \{1, 2, \dots, 1000\}$ maps the contents of $\mathcal{X}$ to ImageNet labels and $H(l(\mathcal{X}_c))$ computes the entropy of the ImageNet label distribution within $\mathcal{X}_c$\footnote{In cluster analysis, entropy is used as an external measure of cluster quality~\cite{karypis2000comparison}.}.
$\Pi(\mathcal{X}_c)=1$ means that all the images in $\mathcal{X}_c$ share the same ImageNet label. 
Since in this case this label has been manually provided, high purity strongly correlates with high interpretability.  

\Cref{t:sela-coherence} reports the coherence for self-supervised classes over different purity ranges.
Within each range, we sample 10 different classes %
for evaluation on AMT and report their average coherence.
In the experiments using random negatives, high purity translates to high semantic coherence for both SeLa and MoCo, while very low purity generally corresponds to low coherence.
We have observed that often such classes consist of bad quality images (\eg blurry or grainy).
Surprisingly, coherence shows a sharp increase with growing purity, also noticeable in~\cref{fig:all_clusters}.
An interesting observation is that most of the classes of intermediate purity (0.5--0.8) appear to be highly coherent.
This suggests that there exist self-supervised classes which are found to be ``interpretable'' yet do not align naturally with an ImageNet label. 
Some examples of such classes are shown in the Appendix. 

On the other hand, using hard negatives we pose a stricter question, \ie whether there are sufficient fine-grained differences between similar classes to make them learnable.
It appears that this is often not the case, as suggested by the significant drop in coherence in \Cref{t:sela-coherence}. 
This is unsurprising given that the methods we analyse find a large number of clusters ($3\times$ the number of labels in ImageNet) and thus likely over-fragment the data. 
It also indicates that, while clusters are often semantically coherent, they are not necessarily ``complete'', in the sense of encompassing all the images that should be grouped together. Finding the right number of clusters remains an open challenge in literature. %

Although SeLa and MoCo are conceptually different methods, our assessment shows similar learnability scores for the resulting clusters, with MoCo showing a slight advantage in the lowest purity group with random negatives and in the highest for hard. 
To test whether the clusters themselves are similar, we compute their adjusted normalized mutual information (aNMI) against ImageNet, a standard metric to estimate similarity between clusterings. 
We obtain an aNMI of $40.7$ for SeLa and $37.3$ for MoCo. %
However, comparing SeLa to MoCo results in a higher aNMI of $44.0$, which means the clusterings are more similar to each other than to ImageNet, despite their methodological differences.
We also observe visual similarities qualitatively and present some examples in the Appendix. 

\begin{figure}[t]
    \centering
    \includegraphics[width=\linewidth]{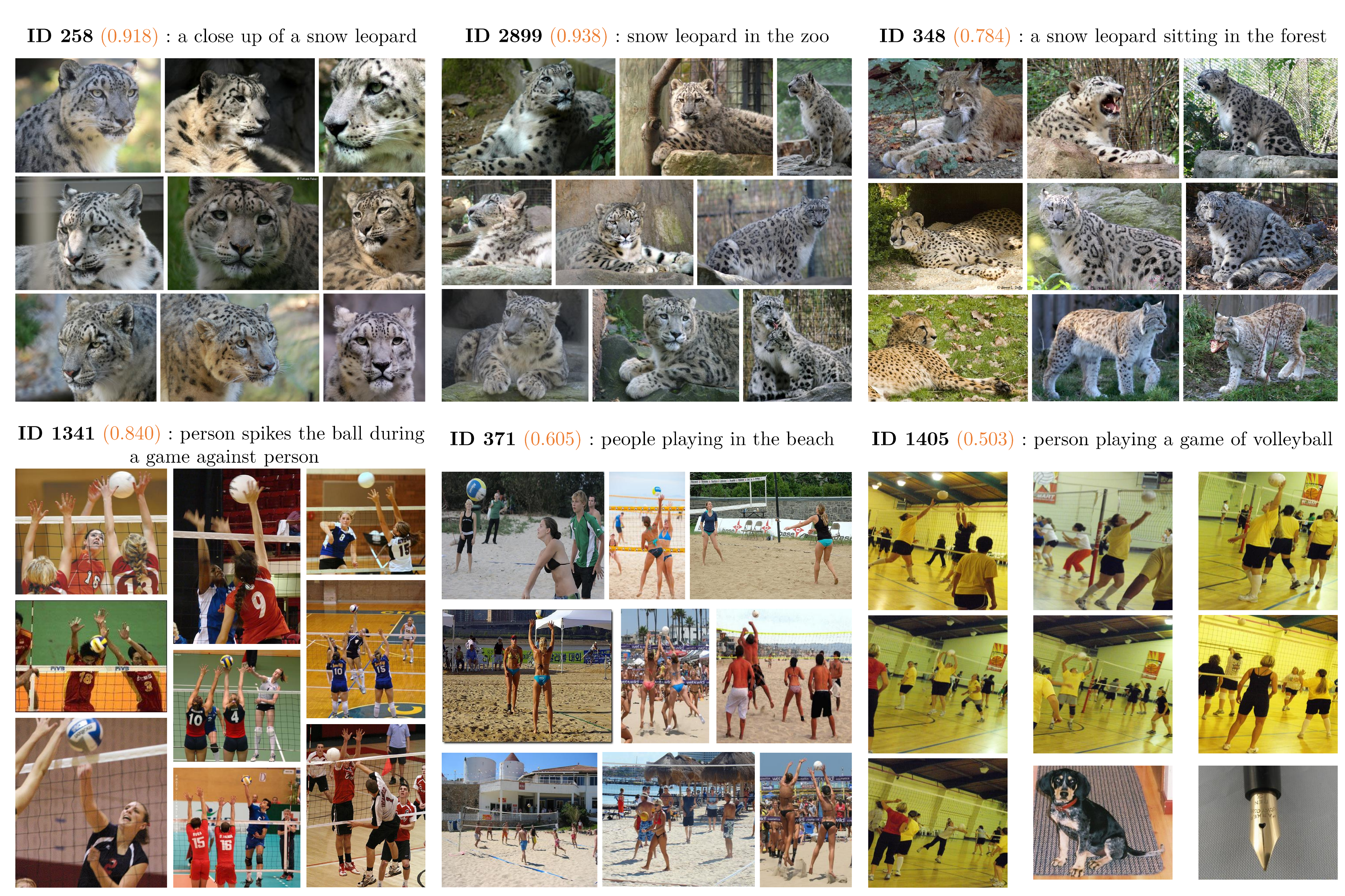}
    \caption{Visualization of SeLa~\cite{asano20self-labelling}-discovered classes with shared concepts (as defined by ImageNet labels); top: \emph{snow leopard}; bottom: \emph{volleyball}. Despite sharing a common concept, the cluster differences are easily recognisable by humans and are even reflected in the automatic descriptions.}\label{fig:similar_clusters}
\end{figure}

\paragraph{Describability.}
Being coherent is not the same as being describable: the latter also requires that the ``gist'' of the class can be described succinctly in language.
Next, we assess the describability of the above clusters, using manually written class descriptions.
Learnability and describability are clearly correlated. 
While describability is generally slightly lower than coherence, sometimes we observe higher describability than coherence (\Cref{t:sela-describability}).
This occurs for classes where a description  (\eg ``low quality photo'') is more explicit and effective than a few visual examples in characterising the class.
Still, lower coherence usually makes a class  $\mathcal{X}_c$ harder to describe as a whole, and as a result the description might only capture a subset of $\mathcal{X}_c$. 

\subsection{Automatic class-level descriptions}

\paragraph{Implementation details.}
To assess the describability of self-supervised classes with automatically generated descriptions, we first obtain captions for individual images using an off-the-shelf captioning model inspired by~\cite{rennie2017self} (Att2in).
We use a publicly available implementation\footnote{\url{https://github.com/ruotianluo/GoogleConceptualCaptioning}} of the model trained on the Conceptual Captions dataset~\cite{sharma2018conceptual}.
We then extract 1024-dimensional caption embeddings using Sentence BERT~\cite{reimers2019sentence} (with \texttt{BERT-large} as the backbone)\footnote{\url{https://github.com/UKPLab/sentence-transformers}} trained on data from the Semantic Textual Similarity benchmark~\cite{cer2018universal}, such that similar captions are represented by similar embeddings.
\Cref{eq:cap_opt} yields the most representative caption for the class, which we use as the class description.

\paragraph{Evaluation.}
Our results on describability with automatic class descriptions are also shown in \cref{t:sela-describability} and \cref{fig:all_clusters}.
Our findings show that the automatic system yields adequate descriptions for visual groups, with only a small gap to the respective manual/expert descriptions.
In~\cref{fig:similar_clusters}, we further show an interesting case of classes from \cite{asano20self-labelling}, where each row shares a visual concept, according to ImageNet annotations.
Nevertheless, differences between these classes are indeed observable, \eg portrait photos vs.~full-body views of the leopard or the difference in the background environment. 
We also notice the distinction between indoor and \emph{beach} volleyball, which is not a label in ImageNet.
Importantly, we note that these differences are also captured by the automatic class description, also shown in the figure.
Further results are shown in the Appendix. 

In addition, we propose an automated method to approximate describability and use it to compare our approach to various baselines.
In \cref{t:google_search}, given a description for an unsupervised class, we retrieve $N=10$ external images using a search engine (\eg Google Image Search) and test whether these are classified in the same way by the unsupervised method (reporting Top-1 or Top-5 accuracy), thus weakly testing the quality of the description.
We report this for all SeLa classes and for the ones used in the AMT experiments (computing recall: R@1, R@5).
In order for this metric to be a proxy for our experiments with humans (\ie binary tests), we also compute it for binary comparisons. 
For each image retrieved for $\mathcal{X}_c$ based on $D_c$, we randomly sample a negative among the images retrieved for other classes.  
We then compare their respective probabilities for class $c$ and report the percentage of correct outcomes, \ie when the positive example is preferred.

Using this metric, in~\cref{t:google_search} we ablate our class-level descriptor generator and compare it to alternatives descriptions.
First, dropping the negative term from \cref{eq:cap_opt}(w/o neg) results in a performance drop in all metrics; this indicates the importance of encouraging distinctive descriptions by contrasting against the other classes. 
Second, we compare to using ROUGE-L~\cite{lin2004rouge} as the distance function in \cref{eq:cap_opt}, thus carrying out the optimization directly on word level. 
This results in a larger performance drop, highlighting the effectiveness of the sentence embedding space in extracting representative descriptions. 
Lastly, we compare to reverse image search, \ie using an image as the search query.
We experiment with both sampling a query image at random from a class and selecting the most representative image based on ResNet-50 features.
In both cases, reverse image search yields significantly worse performance.
Overall, we demonstrate our proposed method is a strong baseline on our describability assessments and also the closest to the results obtained using manual descriptions.

\begin{table}[t]
\centering
\footnotesize
\caption{Classification accuracy (\%) of retrieved images (Google search). For binary accuracy we report the mean and standard deviation over 10 runs. We compare our approach to various baselines.}
\label{t:google_search}
\vspace{0.5em}
\small
\begin{tabular}[t]{l ccc@{\hskip 20pt}cc}
\toprule
& \multicolumn{3}{@{}c}{All classes} & \multicolumn{2}{@{}c}{Selected classes}\\
\cmidrule(lr{19pt}){2-4}
\cmidrule(l{-1pt}r{1pt}){5-6}
Method & Top-1 & Top-5 & Binary & R@1 & R@5 \\ %
\midrule
Random Image & 4.83 & 12.83 & 82.50 $\pm$ 0.14 & 7.16 & 17.62 \\
Representative Image & 4.05 & 17.41 & 86.00 $\pm$ 0.15 & 12.48 & 27.74 \\
\midrule
ROUGE~\cite{lin2004rouge} & 6.12 & 16.06 & 87.77 $\pm$ 0.09 & 12.18 & 28.18\\
$\mathcal{S}_c$ (w/o neg) & 7.64 & 19.10 & 89.48 $\pm$ 0.07 & 11.71 & 27.14 \\
$\mathcal{S}_c\,$ \& $\,\mathcal{S} - \mathcal{S}_c$ & \textbf{7.88} & \textbf{20.29} & \textbf{91.18 $\pm$ 0.07} & \textbf{12.57} & \textbf{30.21} \\
\midrule
Manual description & - & - & - & 18.35 & 34.30 \\ %
\bottomrule
\end{tabular}

\end{table}

\section{Conclusions}\label{s:conclusions}
We have presented two novel concepts, visual learnability and describability, for quantifying the quality of learned image groupings.
With rapidly improving performance of self-supervised methods, this quantification has three implications.
First, understanding, describing and analyzing self-learned categories is an important direction in interpretability research.
Second, when moving away from labelled data, it becomes necessary to evaluate methods without ground truth annotations.
Finally, understanding the difference between human learnability and automatic class discovery can potentially lead to developing improved methods in the future. 

\clearpage\section*{Broader Impact}
\paragraph{Interpretability tools for understanding feature representations.}
Recently, a number of works have focused on explaining or interpreting deep learning models; such research is often known as explainable AI (XAI) or interpretability~\cite{gilpin2018explaining}.
Due to the highly-parameterized nature of CNNs, most researchers treat such models as black-boxes and primarily evaluate them based on task performance on well-curated datasets (\eg ImageNet classification).
However, as deep learning is increasingly applied to high-impact, yet high-risk domains (\eg autonomous driving and medical applications), there is a great need for tools that help us understand CNNs, so that we can in turn understand their limitations and biases.
Our work contributes to the development of interpretability tools that can help society to responsible use and interrogate advanced technology built on deep learning.
We primarily do this in two ways.

\paragraph{Development of principled interpretability metrics.}
First, we present a principled framework for evaluating the human-interpretability of CNN representations.
While this may seem trivial, the interpretability research community has been lagging behind in the development of such metrics.
There have been two main shortcomings of most interpretability evaluation: 1., they are often based on subjective or qualitative inspection, and 2., they fail to evaluate the faithfulness and interpretability of an explanation, that is, it should be both an accurate description of CNN behavior and easy-to-understand.
These two shortcomings often go hand-in-hand.
For example,~\cite{mahendran16salient,adebayo18sanity,fong2019explanations,rebuffi2020there} highlight this issue for attribution heatmaps, which explain what parts of an image are responsible for the model's output decision.
In particular,~\cite{adebayo18sanity} shows that a number of attribution methods that are typically preferred for their visual appearance actually do not accurately describe the CNN being explained. Most metrics focus on evaluating the interpretability of an explanation without also measuring its faithfulness.
This is a major limitation, as an explanation is not useful if it does not accurately describe the phenomenon being explained.

The typical methodology for human evaluation of CNN interpretability asks humans subjective questions like, ``which explanatory visualization do you prefer or trust more?''~\cite{zhou2018interpretable}, ``do these images systematically describe a common visual concept?''~\cite{gonzalez2018semantic}, and ``if so, name that concept''~\cite{zhou2018interpreting}.
Such evaluations tend to evaluate the interpretability without faithfulness (\ie how can we verify that this is the most accurate name for the concept?).
In contrast, our work evaluates using both criteria by shifting from using humans as subjective annotators to using them as more learners that can be evaluated objectively.
Our coherence metric objectively measures how interpretable a CNN-discovered cluster of images is, while our describability metric quantifies how faithfully a natural language description accurately characterizes such a cluster.
We hope that our work serves as a springboard for future work that enables the use of human annotators in evaluating the interpretability of CNNs in a more principled manner.

\paragraph{Understanding self-supervised representations.}
Second, we focus on understanding self-supervised representations.
Most work to date has focused on understanding CNNs trained for image classification.\footnote{
This machine learning \href{http://interpretable-ml.org/icml2020workshop}{workshop} highlights this over-emphasis and encourages more diverse XAI work.
}
However, supervised methods like image classifiers are limited in that they require expensive, manual annotation of highly-curated datasets.
Thus, recent developments of self- and un-supervised methods is exciting, as they do not require manual labels.
That said, there has been relatively little work dedicated to understanding self-supervised representations.
The few works that do explore self-supervised representations typically apply techniques developed on supervised image classifiers to them~\cite{bau17network,fong18net2vec}.

In contrast, we developed our evaluation paradigm with self-supervised methods in mind.
In particular, we were motivated to develop an evaluation framework that could measure the interpretability of coherent, visual concepts that fall outside the limits of being described by labelled datasets.
For example, in~\cref{fig:similar_clusters}, we show that one self-supervised method discovered distinct clusters that highlight different environments of the same concept (\eg different environments for playing volleyball).
Standard interpretability methods of describing such clusters using a labelled dataset~\cite{bau17network,fong18net2vec} would likely map them onto the same label (\eg ``volleyball'') and fail to characterize the subtle nuances captured by different clusters.
Lastly, by design, our paradigm is agnostic to method and can also be used to understand other kinds of image representations, including non-CNN ones.
We hope our work encourages further research on understanding other kinds of representations beyond image classifiers and developing interpretability methods explicitly for those settings.

\section*{Acknowledgements and Funding Disclosure}
We would like to thank Yuki Asano and Christian Rupprecht for helpful discussions and for their feedback on this work. 
We are also grateful for the EPSRC programme grant Seebibyte EP/M013774/1 (I.L.), ERC starting grant IDIU 638009 (I.L), and Open Philanthropy Project (R.F.).

\bibliographystyle{plainnat}
\bibliography{refs,vedaldi,vedaldi_general}

\begin{thebibliography}{78}
\providecommand{\natexlab}[1]{#1}
\providecommand{\url}[1]{\texttt{#1}}
\expandafter\ifx\csname urlstyle\endcsname\relax
  \providecommand{\doi}[1]{doi: #1}\else
  \providecommand{\doi}{doi: \begingroup \urlstyle{rm}\Url}\fi

\bibitem[Abbott et~al.(2011)Abbott, Heller, Ghahramani, and
  Griffiths]{abbott2011testing}
Joshua~T Abbott, Katherine~A Heller, Zoubin Ghahramani, and Thomas~L Griffiths.
\newblock Testing a bayesian measure of representativeness using a large image
  database.
\newblock In \emph{Advances in neural information processing systems}, pages
  2321--2329, 2011.

\bibitem[Adebayo et~al.(2018)Adebayo, Gilmer, Muelly, Goodfellow, Hardt, and
  Kim]{adebayo18sanity}
Julius Adebayo, Justin Gilmer, Michael Muelly, Ian~J. Goodfellow, Moritz Hardt,
  and Been Kim.
\newblock Sanity checks for saliency maps.
\newblock In \emph{Proc. {NeurIPS}}, 2018.

\bibitem[Asano et~al.(2020)Asano, Rupprecht, and
  Vedaldi]{asano20self-labelling}
Yuki~M. Asano, Christian Rupprecht, and Andrea Vedaldi.
\newblock Self-labelling via simultaneous clustering and representation
  learning.
\newblock In \emph{Proceedings of the International Conference on Learning
  Representations ({ICLR})}, 2020.

\bibitem[Bachman et~al.(2019)Bachman, Hjelm, and
  Buchwalter]{bachman2019learning}
Philip Bachman, R~Devon Hjelm, and William Buchwalter.
\newblock Learning representations by maximizing mutual information across
  views.
\newblock In \emph{Advances in Neural Information Processing Systems}, pages
  15509--15519, 2019.

\bibitem[Bau et~al.(2017)Bau, Zhou, Khosla, Oliva, and Torralba]{bau17network}
David Bau, Bolei Zhou, Aditya Khosla, Aude Oliva, and Antonio Torralba.
\newblock Network dissection: Quantifying interpretability of deep visual
  representations.
\newblock In \emph{Proc. {CVPR}}, 2017.

\bibitem[Bojanowski and Joulin(2017)]{bojanowski2017unsupervised}
Piotr Bojanowski and Armand Joulin.
\newblock Unsupervised learning by predicting noise.
\newblock In \emph{Proceedings of the 34th International Conference on Machine
  Learning-Volume 70}, pages 517--526, 2017.

\bibitem[Bravo and Nakayama(1992)]{bravo1992role}
Mary~J Bravo and Ken Nakayama.
\newblock The role of attention in different visual-search tasks.
\newblock \emph{Perception \& psychophysics}, 51\penalty0 (5):\penalty0
  465--472, 1992.

\bibitem[Caron et~al.(2018)Caron, Bojanowski, Joulin, and Douze]{caron2018deep}
Mathilde Caron, Piotr Bojanowski, Armand Joulin, and Matthijs Douze.
\newblock Deep clustering for unsupervised learning of visual features.
\newblock In \emph{Proceedings of the European Conference on Computer Vision
  (ECCV)}, pages 132--149, 2018.

\bibitem[Caron et~al.(2019)Caron, Bojanowski, Mairal, and
  Joulin]{caron2019unsupervised}
Mathilde Caron, Piotr Bojanowski, Julien Mairal, and Armand Joulin.
\newblock Unsupervised pre-training of image features on non-curated data.
\newblock In \emph{Proceedings of the International Conference on Computer
  Vision (ICCV)}, 2019.

\bibitem[Caron et~al.(2020)Caron, Misra, Mairal, Goyal, Bojanowski, and
  Joulin]{caron2020unsupervised}
Mathilde Caron, Ishan Misra, Julien Mairal, Priya Goyal, Piotr Bojanowski, and
  Armand Joulin.
\newblock Unsupervised learning of visual features by contrasting cluster
  assignments.
\newblock \emph{arXiv preprint arXiv:2006.09882}, 2020.

\bibitem[Cer et~al.(2018)Cer, Yang, Kong, Hua, Limtiaco, John, Constant,
  Guajardo-Cespedes, Yuan, Tar, et~al.]{cer2018universal}
Daniel Cer, Yinfei Yang, Sheng-yi Kong, Nan Hua, Nicole Limtiaco, Rhomni~St
  John, Noah Constant, Mario Guajardo-Cespedes, Steve Yuan, Chris Tar, et~al.
\newblock Universal sentence encoder.
\newblock \emph{arXiv preprint arXiv:1803.11175}, 2018.

\bibitem[Chen et~al.(2018)Chen, Ji, Sun, Wu, and Su]{chen2018groupcap}
Fuhai Chen, Rongrong Ji, Xiaoshuai Sun, Yongjian Wu, and Jinsong Su.
\newblock Groupcap: Group-based image captioning with structured relevance and
  diversity constraints.
\newblock In \emph{Proceedings of the IEEE conference on computer vision and
  pattern recognition}, pages 1345--1353, 2018.

\bibitem[Chen et~al.(2020)Chen, Kornblith, Norouzi, and Hinton]{chen2020simple}
Ting Chen, Simon Kornblith, Mohammad Norouzi, and Geoffrey Hinton.
\newblock A simple framework for contrastive learning of visual
  representations.
\newblock \emph{arXiv preprint arXiv:2002.05709}, 2020.

\bibitem[Clopper and Pearson(1934)]{clopper1934use}
Charles~J Clopper and Egon~S Pearson.
\newblock The use of confidence or fiducial limits illustrated in the case of
  the binomial.
\newblock \emph{Biometrika}, 26\penalty0 (4):\penalty0 404--413, 1934.

\bibitem[Cui et~al.(2018)Cui, Yang, Veit, Huang, and Belongie]{cui2018learning}
Yin Cui, Guandao Yang, Andreas Veit, Xun Huang, and Serge Belongie.
\newblock Learning to evaluate image captioning.
\newblock In \emph{Proceedings of the IEEE conference on computer vision and
  pattern recognition}, pages 5804--5812, 2018.

\bibitem[Dai and Lin(2017)]{dai2017contrastive}
Bo~Dai and Dahua Lin.
\newblock Contrastive learning for image captioning.
\newblock In \emph{Advances in Neural Information Processing Systems}, pages
  898--907, 2017.

\bibitem[Deng et~al.(2009)Deng, Dong, Socher, Li, Li, and
  Fei-Fei]{deng2009imagenet}
Jia Deng, Wei Dong, Richard Socher, Li-Jia Li, Kai Li, and Li~Fei-Fei.
\newblock Imagenet: A large-scale hierarchical image database.
\newblock In \emph{2009 IEEE conference on computer vision and pattern
  recognition}, pages 248--255. Ieee, 2009.

\bibitem[Doersch et~al.(2015)Doersch, Gupta, and Efros]{doersch15unsupervised}
Carl Doersch, Abhinav Gupta, and Alexei~A. Efros.
\newblock Unsupervised visual representation learning by context prediction.
\newblock In \emph{Proc. {ICCV}}, 2015.

\bibitem[Fong and Vedaldi(2018)]{fong18net2vec}
Ruth Fong and Andrea Vedaldi.
\newblock {Net2Vec}: Quantifying and explaining how concepts are encoded by
  filters in deep neural networks.
\newblock In \emph{Proceedings of the {IEEE} Conference on Computer Vision and
  Pattern Recognition ({CVPR})}, 2018.

\bibitem[Fong and Vedaldi(2019)]{fong2019explanations}
Ruth Fong and Andrea Vedaldi.
\newblock Explanations for attributing deep neural network predictions.
\newblock In \emph{Explainable AI: Interpreting, Explaining and Visualizing
  Deep Learning}, pages 149--167. Springer, 2019.

\bibitem[Forbes et~al.(2019)Forbes, Kaeser-Chen, Sharma, and
  Belongie]{forbes2019neural}
Maxwell Forbes, Christine Kaeser-Chen, Piyush Sharma, and Serge Belongie.
\newblock Neural naturalist: Generating fine-grained image comparisons.
\newblock \emph{arXiv preprint arXiv:1909.04101}, 2019.

\bibitem[Gilpin et~al.(2018)Gilpin, Bau, Yuan, Bajwa, Specter, and
  Kagal]{gilpin2018explaining}
Leilani~H Gilpin, David Bau, Ben~Z Yuan, Ayesha Bajwa, Michael Specter, and
  Lalana Kagal.
\newblock Explaining explanations: An overview of interpretability of machine
  learning.
\newblock In \emph{2018 IEEE 5th International Conference on data science and
  advanced analytics (DSAA)}, pages 80--89. IEEE, 2018.

\bibitem[Gilton et~al.(2020)Gilton, Luo, Willett, and
  Shakhnarovich]{gilton2020detection}
Davis Gilton, Ruotian Luo, Rebecca Willett, and Greg Shakhnarovich.
\newblock Detection and description of change in visual streams.
\newblock \emph{arXiv preprint arXiv:2003.12633}, 2020.

\bibitem[Gonzalez-Garcia et~al.(2018)Gonzalez-Garcia, Modolo, and
  Ferrari]{gonzalez2018semantic}
Abel Gonzalez-Garcia, Davide Modolo, and Vittorio Ferrari.
\newblock Do semantic parts emerge in convolutional neural networks?
\newblock \emph{International Journal of Computer Vision}, 126\penalty0
  (5):\penalty0 476--494, 2018.

\bibitem[Griffiths et~al.(2016)Griffiths, Abbott, and
  Hsu]{griffiths2016exploring}
Thomas~L Griffiths, Joshua~T Abbott, and Anne~S Hsu.
\newblock Exploring human cognition using large image databases.
\newblock \emph{Topics in cognitive science}, 8\penalty0 (3):\penalty0
  569--588, 2016.

\bibitem[He et~al.(2016)He, Zhang, Ren, and Sun]{he2016deep}
Kaiming He, Xiangyu Zhang, Shaoqing Ren, and Jian Sun.
\newblock Deep residual learning for image recognition.
\newblock In \emph{Proceedings of the IEEE conference on computer vision and
  pattern recognition}, pages 770--778, 2016.

\bibitem[He et~al.(2020)He, Fan, Wu, Xie, and Girshick]{he2020momentum}
Kaiming He, Haoqi Fan, Yuxin Wu, Saining Xie, and Ross Girshick.
\newblock Momentum contrast for unsupervised visual representation learning.
\newblock In \emph{Proceedings of the IEEE/CVF Conference on Computer Vision
  and Pattern Recognition}, pages 9729--9738, 2020.

\bibitem[H{\'e}naff et~al.(2019)H{\'e}naff, Srinivas, De~Fauw, Razavi, Doersch,
  Eslami, and Oord]{henaff2019data}
Olivier~J H{\'e}naff, Aravind Srinivas, Jeffrey De~Fauw, Ali Razavi, Carl
  Doersch, SM~Eslami, and Aaron van~den Oord.
\newblock Data-efficient image recognition with contrastive predictive coding.
\newblock \emph{arXiv preprint arXiv:1905.09272}, 2019.

\bibitem[Hjelm et~al.(2018)Hjelm, Fedorov, Lavoie-Marchildon, Grewal, Bachman,
  Trischler, and Bengio]{hjelm2018learning}
R~Devon Hjelm, Alex Fedorov, Samuel Lavoie-Marchildon, Karan Grewal, Phil
  Bachman, Adam Trischler, and Yoshua Bengio.
\newblock Learning deep representations by mutual information estimation and
  maximization.
\newblock \emph{arXiv preprint arXiv:1808.06670}, 2018.

\bibitem[Hsu et~al.(2012)Hsu, Martin, Sanborn, and
  Griffiths]{hsu2012identifying}
Anne Hsu, Jay Martin, Adam Sanborn, and Tom Griffiths.
\newblock Identifying representations of categories of discrete items using
  markov chain monte carlo with people.
\newblock In \emph{Proceedings of the Annual Meeting of the Cognitive Science
  Society}, volume~34, 2012.

\bibitem[Hu et~al.(2017)Hu, Andreas, Rohrbach, Darrell, and
  Saenko]{hu17learning}
Ronghang Hu, Jacob Andreas, Marcus Rohrbach, Trevor Darrell, and Kate Saenko.
\newblock Learning to reason: End-to-end module networks for visual question
  answering.
\newblock In \emph{Proc. {ICCV}}, 2017.

\bibitem[Jhamtani and Berg{-}Kirkpatrick(2018)]{JhamtaniB18}
Harsh Jhamtani and Taylor Berg{-}Kirkpatrick.
\newblock Learning to describe differences between pairs of similar images.
\newblock In \emph{Proceedings of the 2018 Conference on Empirical Methods in
  Natural Language Processing, Brussels, Belgium, October 31 - November 4,
  2018}, pages 4024--4034. Association for Computational Linguistics, 2018.

\bibitem[Ji et~al.(2019)Ji, Henriques, and Vedaldi]{ji2019invariant}
Xu~Ji, Jo{\~a}o~F Henriques, and Andrea Vedaldi.
\newblock Invariant information clustering for unsupervised image
  classification and segmentation.
\newblock In \emph{Proceedings of the IEEE International Conference on Computer
  Vision}, pages 9865--9874, 2019.

\bibitem[Jia et~al.(2013)Jia, Abbott, Austerweil, Griffiths, and
  Darrell]{jia2013visual}
Yangqing Jia, Joshua~T Abbott, Joseph~L Austerweil, Tom Griffiths, and Trevor
  Darrell.
\newblock Visual concept learning: Combining machine vision and bayesian
  generalization on concept hierarchies.
\newblock In \emph{Advances in Neural Information Processing Systems}, pages
  1842--1850, 2013.

\bibitem[Jing and Tian(2020)]{jing2020self}
Longlong Jing and Yingli Tian.
\newblock Self-supervised visual feature learning with deep neural networks: A
  survey.
\newblock \emph{IEEE Transactions on Pattern Analysis and Machine
  Intelligence}, 2020.

\bibitem[Karypis and Kumar()]{karypis2000comparison}
Michael Steinbach~George Karypis and Vipin Kumar.
\newblock A comparison of document clustering techniques.

\bibitem[Kilickaya et~al.(2017)Kilickaya, Erdem, Ikizler{-}Cinbis, and
  Erdem]{KilickayaEIE17}
Mert Kilickaya, Aykut Erdem, Nazli Ikizler{-}Cinbis, and Erkut Erdem.
\newblock Re-evaluating automatic metrics for image captioning.
\newblock In \emph{Proceedings of the 15th Conference of the European Chapter
  of the Association for Computational Linguistics, {EACL} 2017, Valencia,
  Spain, April 3-7, 2017, Volume 1: Long Papers}, pages 199--209. Association
  for Computational Linguistics, 2017.

\bibitem[Kim et~al.(2017)Kim, Wattenberg, Gilmer, Cai, Wexler, Vi{\'{e}}gas,
  and Sayres]{kim17interpretability}
Been Kim, Martin Wattenberg, Justin Gilmer, Carrie~J. Cai, James Wexler,
  Fernanda~B. Vi{\'{e}}gas, and Rory Sayres.
\newblock Interpretability beyond feature attribution: Quantitative testing
  with concept activation vectors (tcav).
\newblock In \emph{Proc. {ICML}}, 2017.

\bibitem[Kiros et~al.(2015)Kiros, Zhu, Salakhutdinov, Zemel, Urtasun, Torralba,
  and Fidler]{kiros2015skip}
Ryan Kiros, Yukun Zhu, Russ~R Salakhutdinov, Richard Zemel, Raquel Urtasun,
  Antonio Torralba, and Sanja Fidler.
\newblock Skip-thought vectors.
\newblock In \emph{Advances in neural information processing systems}, pages
  3294--3302, 2015.

\bibitem[Krippendorff(2011)]{krippendorff2011computing}
Klaus Krippendorff.
\newblock Computing krippendorff's alpha-reliability.
\newblock 2011.

\bibitem[Li et~al.(2020)Li, Tran, Mai, Lin, and Yuille]{li2020context}
Zhuowan Li, Quan Tran, Long Mai, Zhe Lin, and Alan~L Yuille.
\newblock Context-aware group captioning via self-attention and contrastive
  features.
\newblock In \emph{Proceedings of the IEEE/CVF Conference on Computer Vision
  and Pattern Recognition}, pages 3440--3450, 2020.

\bibitem[Lin(2004)]{lin2004rouge}
Chin-Yew Lin.
\newblock Rouge: A package for automatic evaluation of summaries.
\newblock In \emph{Text Summarization Branches Out}, pages 74--81, 2004.

\bibitem[Liu et~al.(2018)Liu, Li, Shao, Chen, and Wang]{liu2018show}
Xihui Liu, Hongsheng Li, Jing Shao, Dapeng Chen, and Xiaogang Wang.
\newblock Show, tell and discriminate: Image captioning by self-retrieval with
  partially labeled data.
\newblock In \emph{Proceedings of the European Conference on Computer Vision
  (ECCV)}, pages 338--354, 2018.

\bibitem[Luo et~al.(2018)Luo, Price, Cohen, and
  Shakhnarovich]{luo2018discriminability}
Ruotian Luo, Brian Price, Scott Cohen, and Gregory Shakhnarovich.
\newblock Discriminability objective for training descriptive captions.
\newblock In \emph{Proceedings of the IEEE Conference on Computer Vision and
  Pattern Recognition}, pages 6964--6974, 2018.

\bibitem[Mahendran and Vedaldi(2016{\natexlab{a}})]{mahendran16salient}
Aravindh Mahendran and Andrea Vedaldi.
\newblock Salient deconvolutional networks.
\newblock In \emph{Proceedings of the European Conference on Computer Vision
  ({ECCV})}, 2016{\natexlab{a}}.

\bibitem[Mahendran and Vedaldi(2016{\natexlab{b}})]{mahendran16visualizing}
Aravindh Mahendran and Andrea Vedaldi.
\newblock Visualizing deep convolutional neural networks using natural
  pre-images.
\newblock \emph{International Journal of Computer Vision ({IJCV})},
  120\penalty0 (3), 2016{\natexlab{b}}.

\bibitem[Mihalcea and Tarau(2004)]{mihalcea2004textrank}
Rada Mihalcea and Paul Tarau.
\newblock {T}ext{R}ank: Bringing order into text.
\newblock In \emph{Proceedings of the 2004 Conference on Empirical Methods in
  Natural Language Processing}, pages 404--411, Barcelona, Spain, July 2004.
  Association for Computational Linguistics.

\bibitem[Misra and Maaten(2020)]{misra2020self}
Ishan Misra and Laurens van~der Maaten.
\newblock Self-supervised learning of pretext-invariant representations.
\newblock In \emph{Proceedings of the IEEE/CVF Conference on Computer Vision
  and Pattern Recognition}, pages 6707--6717, 2020.

\bibitem[Nenkova and McKeown(2012)]{nenkova2012survey}
Ani Nenkova and Kathleen McKeown.
\newblock A survey of text summarization techniques.
\newblock In \emph{Mining text data}, pages 43--76. Springer, 2012.

\bibitem[Nguyen et~al.(2017)Nguyen, Clune, Bengio, Dosovitskiy, and
  Yosinski]{nguyen2017plug}
Anh Nguyen, Jeff Clune, Yoshua Bengio, Alexey Dosovitskiy, and Jason Yosinski.
\newblock Plug \& play generative networks: Conditional iterative generation of
  images in latent space.
\newblock In \emph{Proceedings of the IEEE Conference on Computer Vision and
  Pattern Recognition}, pages 4467--4477, 2017.

\bibitem[Nguyen et~al.(2016)Nguyen, Dosovitskiy, Yosinski, Brox, and
  Clune]{nguyen16synthesizing}
Anh~Mai Nguyen, Alexey Dosovitskiy, Jason Yosinski, Thomas Brox, and Jeff
  Clune.
\newblock Synthesizing the preferred inputs for neurons in neural networks via
  deep generator networks.
\newblock In \emph{Proc. {NeurIPS}}, 2016.

\bibitem[Noroozi and Favaro(2016)]{noroozi16unsupervised}
Mehdi Noroozi and Paolo Favaro.
\newblock Unsupervised learning of visual representations by solving jigsaw
  puzzles.
\newblock In \emph{Proc. {ECCV}}, 2016.

\bibitem[Noroozi et~al.(2018)Noroozi, Vinjimoor, Favaro, and
  Pirsiavash]{noroozi2018boosting}
Mehdi Noroozi, Ananth Vinjimoor, Paolo Favaro, and Hamed Pirsiavash.
\newblock Boosting self-supervised learning via knowledge transfer.
\newblock In \emph{Proceedings of the IEEE Conference on Computer Vision and
  Pattern Recognition}, pages 9359--9367, 2018.

\bibitem[Olah et~al.(2017)Olah, Mordvintsev, and Schubert]{olah17feature}
Chris Olah, Alexander Mordvintsev, and Ludwig Schubert.
\newblock Feature visualization.
\newblock \emph{Distill}, 2\penalty0 (11), 2017.

\bibitem[Oord et~al.(2018)Oord, Li, and Vinyals]{oord2018representation}
Aaron van~den Oord, Yazhe Li, and Oriol Vinyals.
\newblock Representation learning with contrastive predictive coding.
\newblock \emph{arXiv preprint arXiv:1807.03748}, 2018.

\bibitem[Park et~al.(2019)Park, Darrell, and Rohrbach]{park2019robust}
Dong~Huk Park, Trevor Darrell, and Anna Rohrbach.
\newblock Robust change captioning.
\newblock In \emph{Proceedings of the IEEE International Conference on Computer
  Vision}, pages 4624--4633, 2019.

\bibitem[Pathak et~al.(2016)Pathak, Krahenbuhl, Donahue, Darrell, and
  Efros]{pathak2016context}
Deepak Pathak, Philipp Krahenbuhl, Jeff Donahue, Trevor Darrell, and Alexei~A
  Efros.
\newblock Context encoders: Feature learning by inpainting.
\newblock In \emph{Proceedings of the IEEE conference on computer vision and
  pattern recognition}, pages 2536--2544, 2016.

\bibitem[Rebuffi et~al.(2020)Rebuffi, Fong, Ji, and Vedaldi]{rebuffi2020there}
Sylvestre-Alvise Rebuffi, Ruth Fong, Xu~Ji, and Andrea Vedaldi.
\newblock There and back again: Revisiting backpropagation saliency methods.
\newblock \emph{arXiv preprint arXiv:2004.02866}, 2020.

\bibitem[Reimers and Gurevych(2019)]{reimers2019sentence}
Nils Reimers and Iryna Gurevych.
\newblock Sentence-bert: Sentence embeddings using siamese bert-networks.
\newblock In \emph{Proceedings of the 2019 Conference on Empirical Methods in
  Natural Language Processing}. Association for Computational Linguistics, 11
  2019.

\bibitem[Rennie et~al.(2017)Rennie, Marcheret, Mroueh, Ross, and
  Goel]{rennie2017self}
Steven~J Rennie, Etienne Marcheret, Youssef Mroueh, Jerret Ross, and Vaibhava
  Goel.
\newblock Self-critical sequence training for image captioning.
\newblock In \emph{Proceedings of the IEEE Conference on Computer Vision and
  Pattern Recognition}, pages 7008--7024, 2017.

\bibitem[Sharma et~al.(2018)Sharma, Ding, Goodman, and
  Soricut]{sharma2018conceptual}
Piyush Sharma, Nan Ding, Sebastian Goodman, and Radu Soricut.
\newblock Conceptual captions: A cleaned, hypernymed, image alt-text dataset
  for automatic image captioning.
\newblock In \emph{Proceedings of the 56th Annual Meeting of the Association
  for Computational Linguistics (Volume 1: Long Papers)}, pages 2556--2565,
  2018.

\bibitem[Simonyan et~al.(2014)Simonyan, Vedaldi, and Zisserman]{simonyan14deep}
Karen Simonyan, Andrea Vedaldi, and Andrew Zisserman.
\newblock Deep inside convolutional networks: Visualising image classification
  models and saliency maps.
\newblock In \emph{Proceedings of the International Conference on Learning
  Representations ({ICLR})}, 2014.

\bibitem[Srinivas et~al.(2020)Srinivas, Laskin, and Abbeel]{srinivas2020curl}
Aravind Srinivas, Michael Laskin, and Pieter Abbeel.
\newblock Curl: Contrastive unsupervised representations for reinforcement
  learning.
\newblock \emph{arXiv preprint arXiv:2004.04136}, 2020.

\bibitem[Tenenbaum et~al.(2001)Tenenbaum, Griffiths,
  et~al.]{tenenbaum2001rational}
Joshua~B Tenenbaum, Thomas~L Griffiths, et~al.
\newblock The rational basis of representativeness.
\newblock pages 1036--1041, 2001.

\bibitem[Tian et~al.(2019)Tian, Krishnan, and Isola]{tian2019contrastive}
Yonglong Tian, Dilip Krishnan, and Phillip Isola.
\newblock Contrastive multiview coding.
\newblock \emph{arXiv preprint arXiv:1906.05849}, 2019.

\bibitem[Ulyanov et~al.(2018)Ulyanov, Vedaldi, and Lempitsky]{ulyanov18deep}
Dmitry Ulyanov, Andrea Vedaldi, and Victor~S. Lempitsky.
\newblock Deep image prior.
\newblock In \emph{Proceedings of the {IEEE} Conference on Computer Vision and
  Pattern Recognition ({CVPR})}, 2018.

\bibitem[Vedantam et~al.(2017)Vedantam, Bengio, Murphy, Parikh, and
  Chechik]{vedantam2017context}
Ramakrishna Vedantam, Samy Bengio, Kevin Murphy, Devi Parikh, and Gal Chechik.
\newblock Context-aware captions from context-agnostic supervision.
\newblock In \emph{Proceedings of the IEEE Conference on Computer Vision and
  Pattern Recognition}, pages 251--260, 2017.

\bibitem[Wu et~al.(2018)Wu, Xiong, Yu, and Lin]{wu2018unsupervised}
Zhirong Wu, Yuanjun Xiong, Stella~X Yu, and Dahua Lin.
\newblock Unsupervised feature learning via non-parametric instance
  discrimination.
\newblock In \emph{Proceedings of the IEEE Conference on Computer Vision and
  Pattern Recognition}, pages 3733--3742, 2018.

\bibitem[Xu and Tenenbaum(2007)]{xu2007word}
Fei Xu and Joshua~B Tenenbaum.
\newblock Word learning as bayesian inference.
\newblock \emph{Psychological review}, 114\penalty0 (2):\penalty0 245, 2007.

\bibitem[Yan et~al.(2020)Yan, Misra, Gupta, Ghadiyaram, and
  Mahajan]{yan2020cluster}
Xueting Yan, Ishan Misra, Abhinav Gupta, Deepti Ghadiyaram, and Dhruv Mahajan.
\newblock Clusterfit: Improving generalization of visual representations.
\newblock In \emph{CVPR}, 2020.

\bibitem[Yang et~al.(2017)Yang, Fu, Sidiropoulos, and Hong]{yang2017towards}
Bo~Yang, Xiao Fu, Nicholas~D Sidiropoulos, and Mingyi Hong.
\newblock Towards k-means-friendly spaces: Simultaneous deep learning and
  clustering.
\newblock In \emph{Proceedings of the 34th International Conference on Machine
  Learning-Volume 70}, pages 3861--3870. JMLR. org, 2017.

\bibitem[Yang et~al.(2016)Yang, Parikh, and Batra]{yang2016joint}
Jianwei Yang, Devi Parikh, and Dhruv Batra.
\newblock Joint unsupervised learning of deep representations and image
  clusters.
\newblock In \emph{Proceedings of the IEEE Conference on Computer Vision and
  Pattern Recognition}, pages 5147--5156, 2016.

\bibitem[Young et~al.(2014)Young, Lai, Hodosh, and Hockenmaier]{young2014image}
Peter Young, Alice Lai, Micah Hodosh, and Julia Hockenmaier.
\newblock From image descriptions to visual denotations: New similarity metrics
  for semantic inference over event descriptions.
\newblock \emph{Transactions of the Association for Computational Linguistics},
  2:\penalty0 67--78, 2014.

\bibitem[Zeiler and Fergus(2014)]{zeiler14visualizing}
Matthew~D. Zeiler and Rob Fergus.
\newblock Visualizing and understanding convolutional networks.
\newblock In \emph{Proc. {ECCV}}, 2014.

\bibitem[Zhou et~al.(2014)Zhou, Khosla, Lapedriza, Oliva, and
  Torralba]{zhou2014object}
Bolei Zhou, Aditya Khosla, Agata Lapedriza, Aude Oliva, and Antonio Torralba.
\newblock Object detectors emerge in deep scene cnns.
\newblock \emph{arXiv preprint arXiv:1412.6856}, 2014.

\bibitem[Zhou et~al.(2018{\natexlab{a}})Zhou, Bau, Oliva, and
  Torralba]{zhou2018interpreting}
Bolei Zhou, David Bau, Aude Oliva, and Antonio Torralba.
\newblock Interpreting deep visual representations via network dissection.
\newblock \emph{IEEE transactions on pattern analysis and machine
  intelligence}, 41\penalty0 (9):\penalty0 2131--2145, 2018{\natexlab{a}}.

\bibitem[Zhou et~al.(2018{\natexlab{b}})Zhou, Sun, Bau, and
  Torralba]{zhou2018interpretable}
Bolei Zhou, Yiyou Sun, David Bau, and Antonio Torralba.
\newblock Interpretable basis decomposition for visual explanation.
\newblock In \emph{Proceedings of the European Conference on Computer Vision
  (ECCV)}, pages 119--134, 2018{\natexlab{b}}.

\bibitem[Zhou and Firestone(2019)]{zhou2019humans}
Zhenglong Zhou and Chaz Firestone.
\newblock Humans can decipher adversarial images.
\newblock \emph{Nature communications}, 10\penalty0 (1):\penalty0 1--9, 2019.

\end{thebibliography}

\newpage
\appendix

\section*{\Large{Appendix}}
\vspace{1.5em}

\section{Qualitative Examples and Discussion}

\begin{figure}
    \centering
    \includegraphics[width=\textwidth]{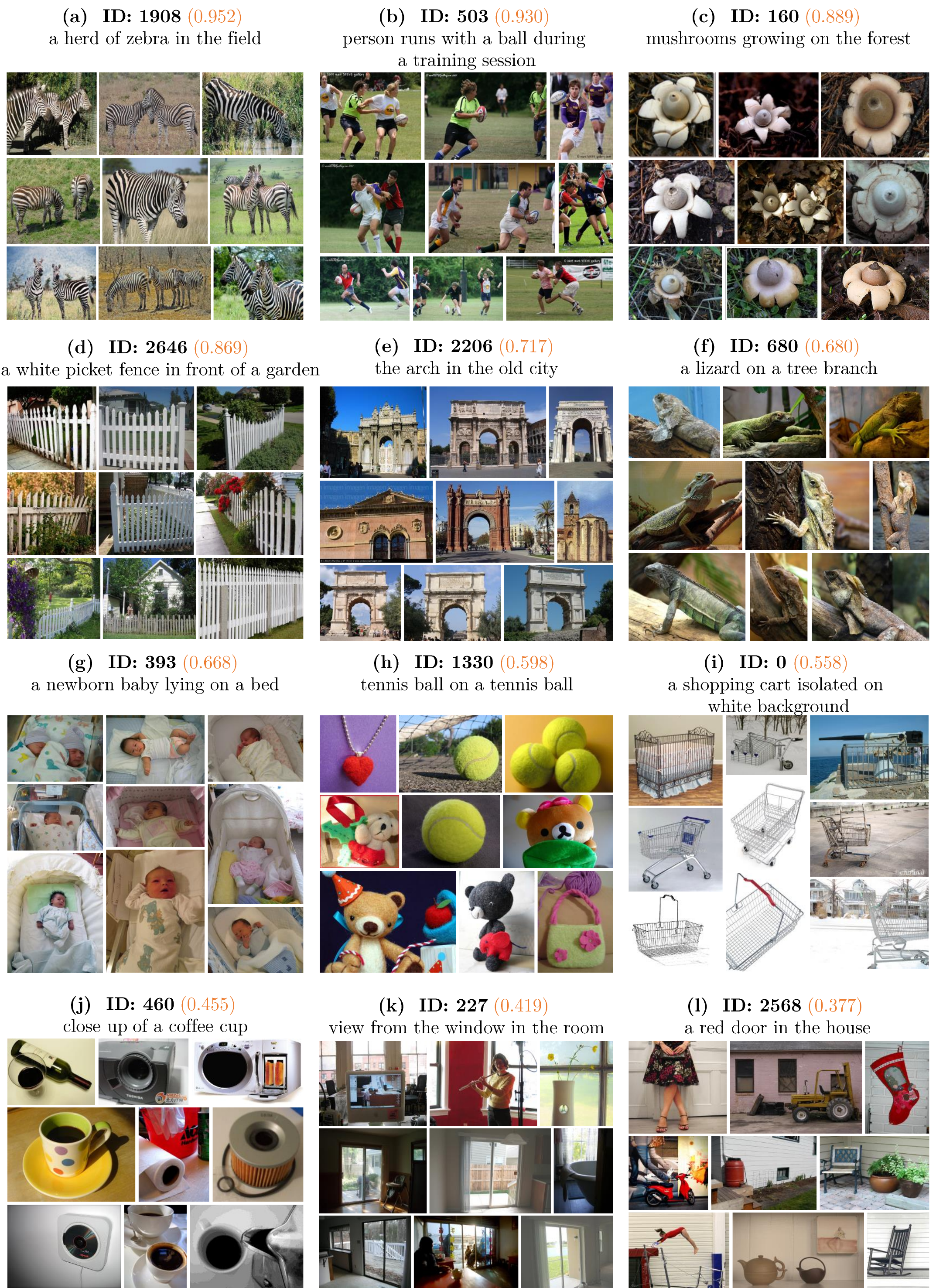}
    \caption{Samples from SeLa classes along with the predicted class captions, sorted by purity (in orange).}
    \label{fig:sela_examples}
\end{figure}

\begin{figure}
    \centering
    \includegraphics[width=\textwidth]{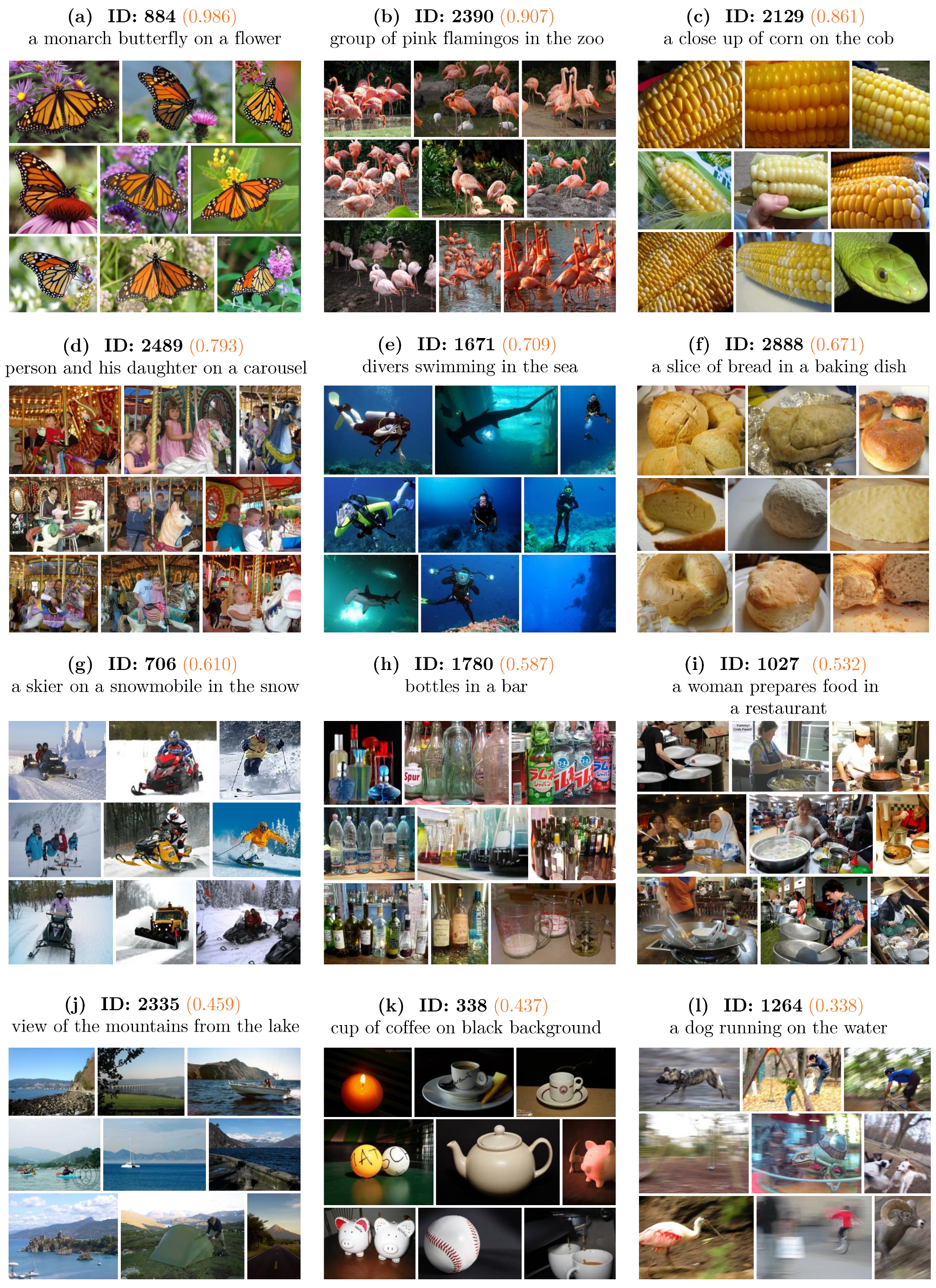}
    \caption{Samples from MoCo classes along with the predicted class captions, sorted by purity (in orange).}
    \label{fig:moco_examples}
\end{figure}

In \cref{fig:sela_examples} and \cref{fig:moco_examples} we show a selection of SeLa and MoCo classes respectively with varying purity (the lowest purity class for both methods has $\Pi \approx 0.3$).
We can make several interesting observations. 

First, we observe that concepts emerging in self-supervised methods, might not necessarily be annotated in ImageNet.
A prominent example is shown in \cref{fig:sela_examples}(g), that is a cluster of \emph{babies}, while the most frequent ImageNet label in this cluster is \emph{bassinet}. 
Notably, while the cluster in question is of seemingly intermediate purity, it exhibits strong semantic coherence and describability, supporting our findings in Table 1 (main paper). 
It is therefore worth asking what is the implication of this on linear classification accuracy on ImageNet (\ie training linear probes) as a way evaluate such methods; we leave this to future work. 

Second, the quality of the class descriptions is often dependent on  cluster quality.
In most cases, \emph{very} low purity clusters have no visual similarities and the concept shared among images (if any) is often difficult to identify, even for humans. For example, \cref{fig:moco_examples}(l) could be \emph{``motion blur in dynamic scene"}. 
In these cases, the automatic class-level description is often unsuccessful in conveying the gist of the class (\eg \cref{fig:sela_examples}(l), \cref{fig:moco_examples}(l)); this is also in agreement with low learnability and describability scores reported in Table 1 (main paper).

Third, we often observe clusters such as the ones shown in \cref{fig:sela_examples}(h), (j), where the prominent visual concepts could perhaps be identified as \emph{``felt material''} and \emph{``black circle''} respectively. 
In such cases, images in the class might be closer visually rather than semantically, \ie sharing material or other visual properties.
While these might be easily recognisable by humans, describing such concepts remains still a challenge for image captioning systems given that existing datasets are highly semantic\,---\,\ie descriptions of textures, patterns, etc. are heavily under-represented. 
As a result, summarising the class by extracting abstract class-level descriptions from image-level semantics is difficult and sometimes results in failure cases. 
For the examples above, the automatic class-level descriptions ``collapse'' to the most prominent semantic concepts in each cluster, \ie \emph{``tennis ball''} and \emph{``cup of coffee''}.
Empirically, we have found that advancing class-level captioning systems will require additional knowledge about the world, for example how individual semantic categories can be linked on different levels, such as material (what are A and B made of?), appearance (how do A and B look like?) or usage (how are A and B used?). 

In \Cref{fig:evaluated_clusters} we show a few of clusters that we evaluated on AMT (20 HITs each), together with their learnability and describability scores. For each case, We also show samples from the hard negative cluster.

\section{Additional Results}

\subsection{Learnability with Hard Negatives}
\begin{figure}
    \centering
    \includegraphics[width=0.95\textwidth]{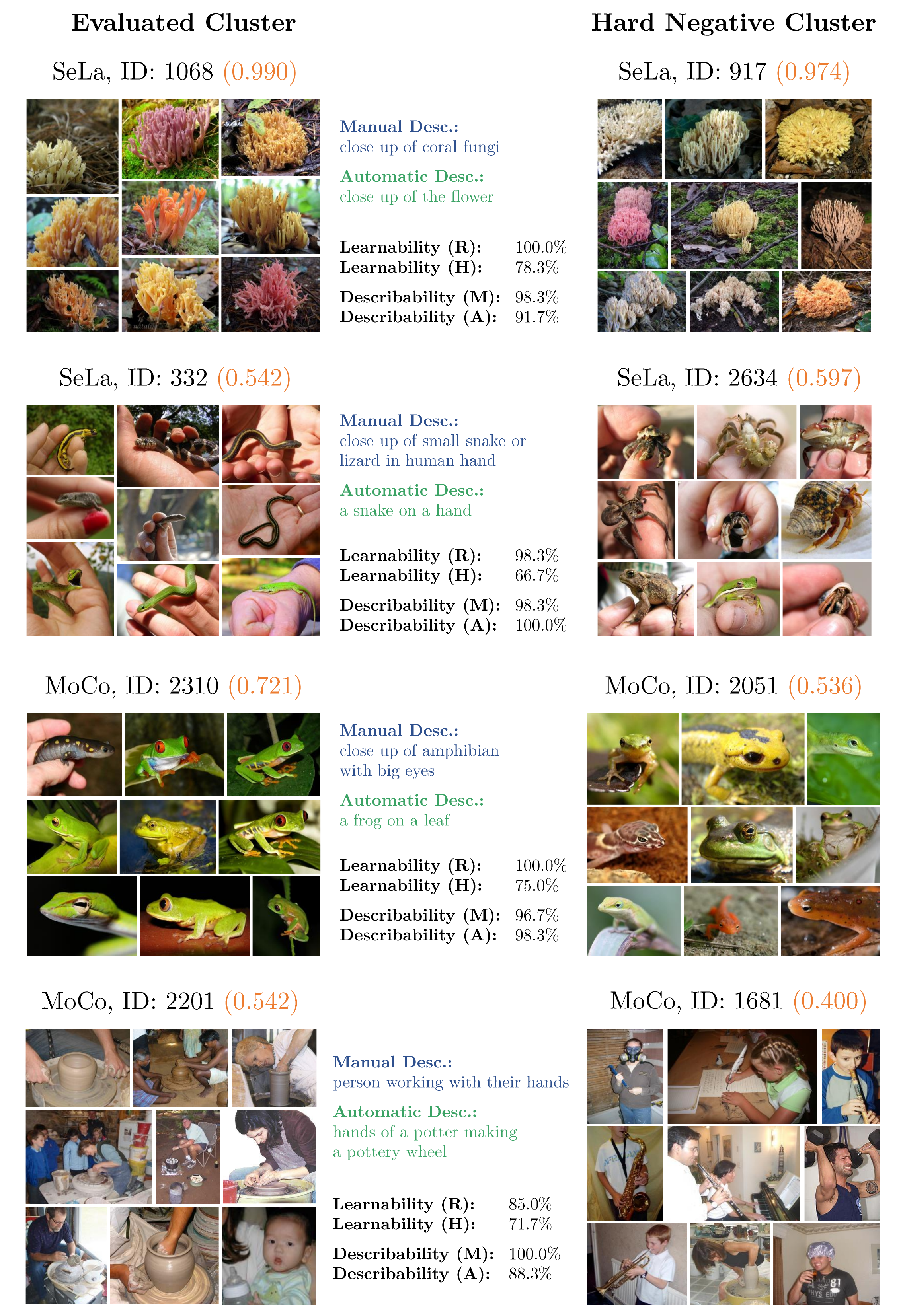}
    \caption{Examples of clusters that were evaluated. On the left, we show the target cluster. The middle column shows the manual (M) and automatic (A) descriptions obtained for the cluster and the evaluation outcome: learnability with random (R) and hard (H) negatives and describability. On the right, we show samples from the hard negative cluster used for the evaluation of learnability (H).}
    \label{fig:evaluated_clusters}
\end{figure}

In addition to \cref{fig:all_clusters} from the main paper, we plot the learnability for each class with hard negative sampling in \cref{fig:HN}. 
Because of over-fragmentation of the data into clusters, the ``hard negative'' counterpart is often a very similar (or even the same) concept, thus resulting in several pure clusters having low learnability scores, despite being highly interpretable. 
Some examples are shown in \cref{fig:evaluated_clusters}
Learnability with hard negatives alone is not a good indicator of the overall coherence of a class. 

\subsection{Comparing Clusterings}
In \cref{t:nmi}, we report the normalized mutual information (NMI), adjusted NMI and adjusted rand index (ARI) to quantify the quality of the clustering against ImageNet. 
We consider two independent clusterings for MoCo, one with 1k and one with 3k classes. 
Interestingly, as discussed in the main paper, we find out that there are more similarities between self-supervised methods SeLa and MoCo than similarities between each method and ImageNet, despite their being fundamentally different approaches.  
This holds true even for MoCo-1k, even though the number of clusters in this case is the same as with ImageNet labels.  

\begin{figure}
    \centering
    \includegraphics[width=\textwidth]{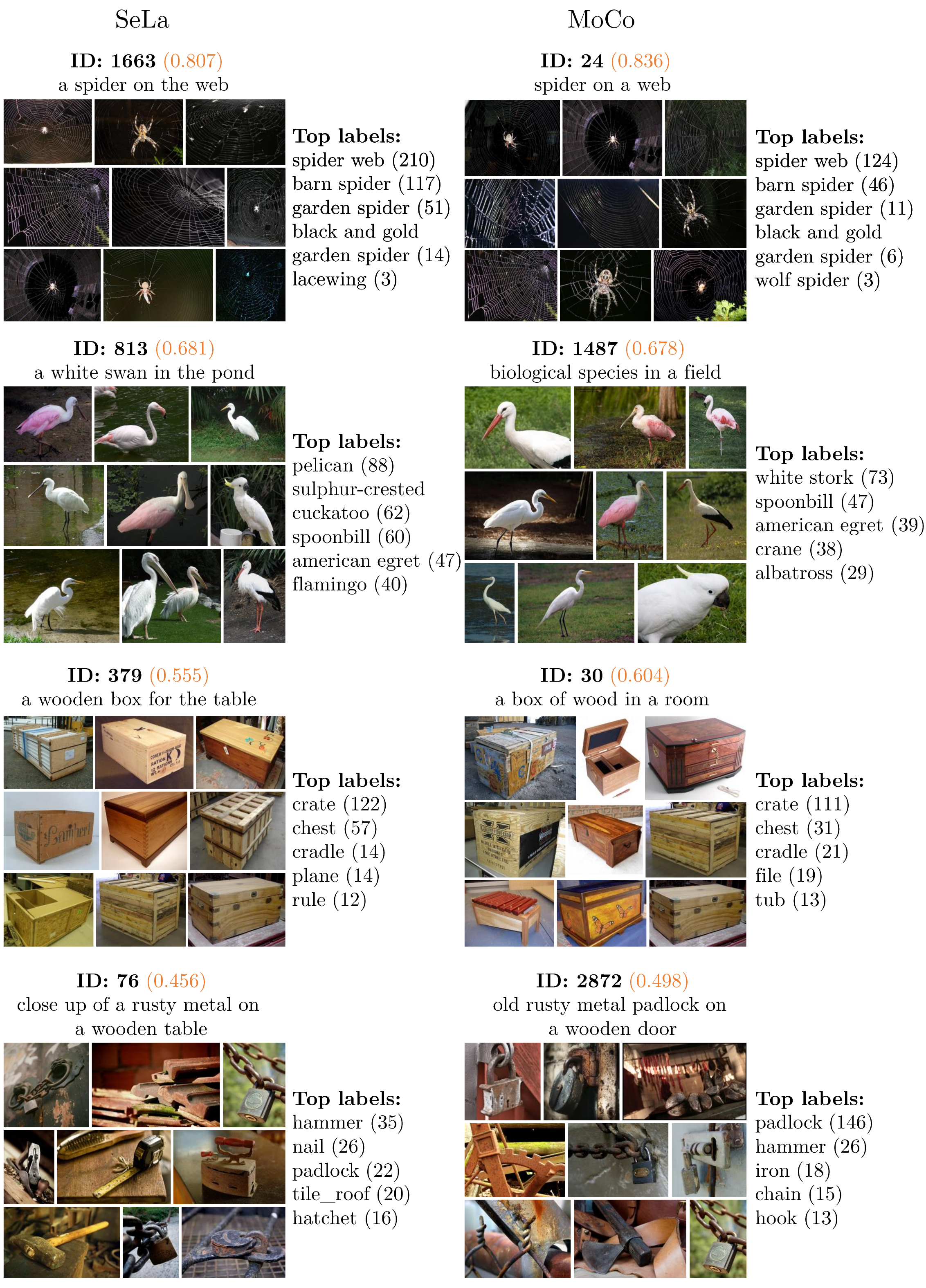}
    \caption{Similar concepts emerging from different self-supervised methods, shown side by side. The top ImageNet labels in each cluster are shown on the right, with the number in parenthesis being the number of occurrences for each category. Notably, corresponding clusters also exhibit similar purity (in orange).}
    \label{fig:sela_moco}
\end{figure}

We can also justify this observation qualitatively, as there exist clusters that are very similar to each other, yet they have intermediate to low purity with respect to ImageNet. 
Some interesting examples are shown side-by-side in \cref{fig:sela_moco}.
This finding suggests that self-supervised approaches discover similar concepts\,---\,which are also \emph{interpretable} by humans\,---\,and that these concepts might not necessarily align with ImageNet labels. 
Notably, in the examples shown in \cref{fig:sela_moco}, similar clusters also have a similar degree of ``impurity'' ($\Pi (\mathcal{X}_c)$ given in orange).  

\begin{table}
	\begin{minipage}{0.48\linewidth}
		\caption{Clustering quality metrics for all permutations of SeLa, MoCo and ImageNet.}
		\vspace{1em}
		\label{t:nmi}
		\centering
		\small
		 \begin{tabular}[t]{l ccc}
        \toprule
        Method & NMI & aNMI & ARI\\
        \midrule
        ImageNet--SeLa & 49.6 & 40.7 & 8.3 \\
        ImageNet--MoCo [1k] & 43.4 & 39.5 & 6.0 \\
        ImageNet--MoCo [3k] & 46.4 & 37.3 & 5.3 \\
        SeLa--MoCo [1k] & 54.2 & 46.4 & 6.8 \\
        SeLa--MoCo [3k] & 58.0 & 44.0 & 9.3 \\
        \bottomrule
        \end{tabular}
    	\end{minipage}\hfill
	    \begin{minipage}{0.49\linewidth}
		\centering
		\includegraphics[width=0.99\linewidth]{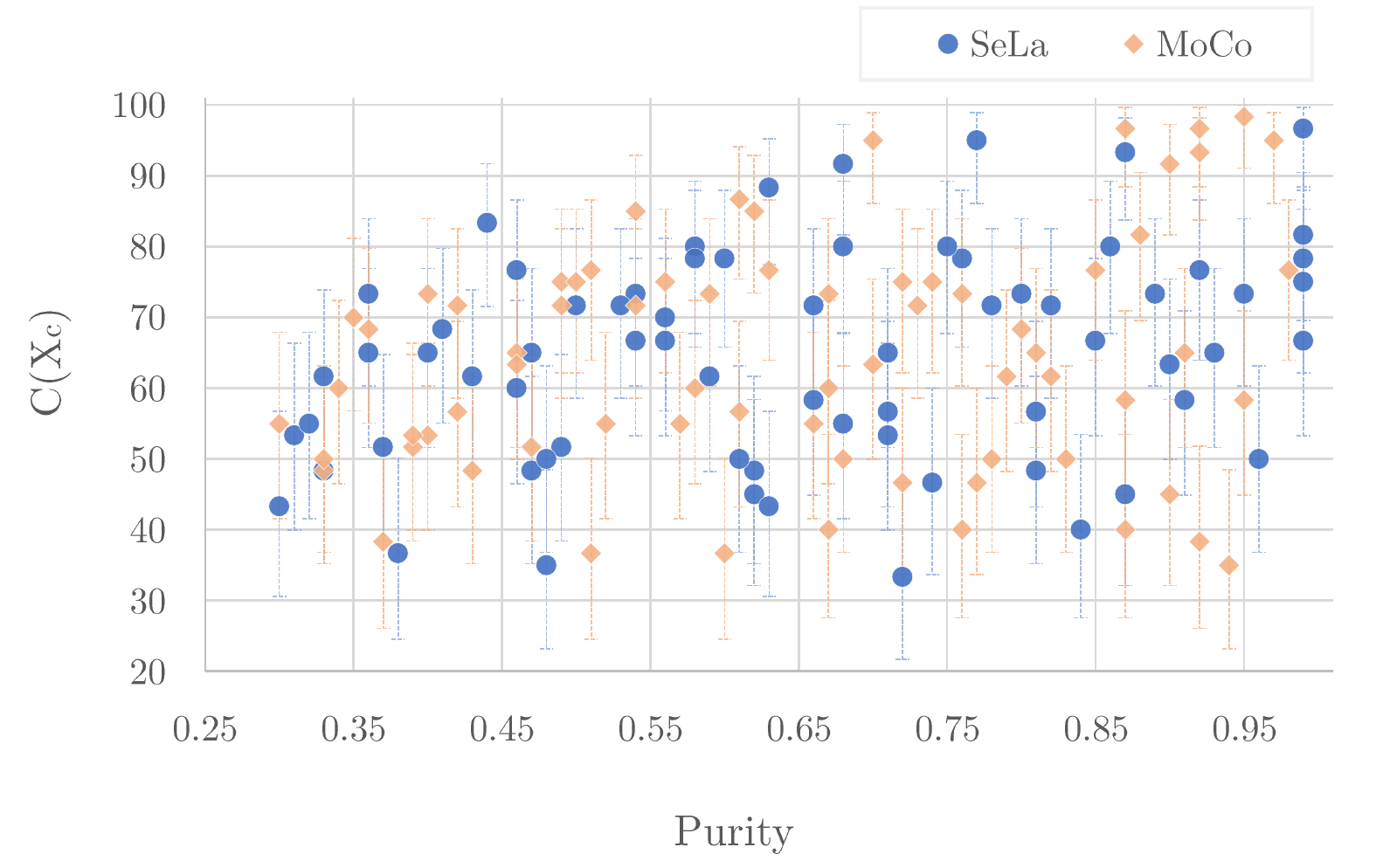}
		\captionof{figure}{Semantic coherence measured with hard negatives (shown per cluster with 95\% CI)}
		\label{fig:HN}
	\end{minipage}
\end{table}

\subsection{Evaluating Caption Quality}
\begin{figure}[t]
\centering
\begin{subcaptionbox}{(a)}[.48\textwidth]
  {\includegraphics[width=.9\linewidth]{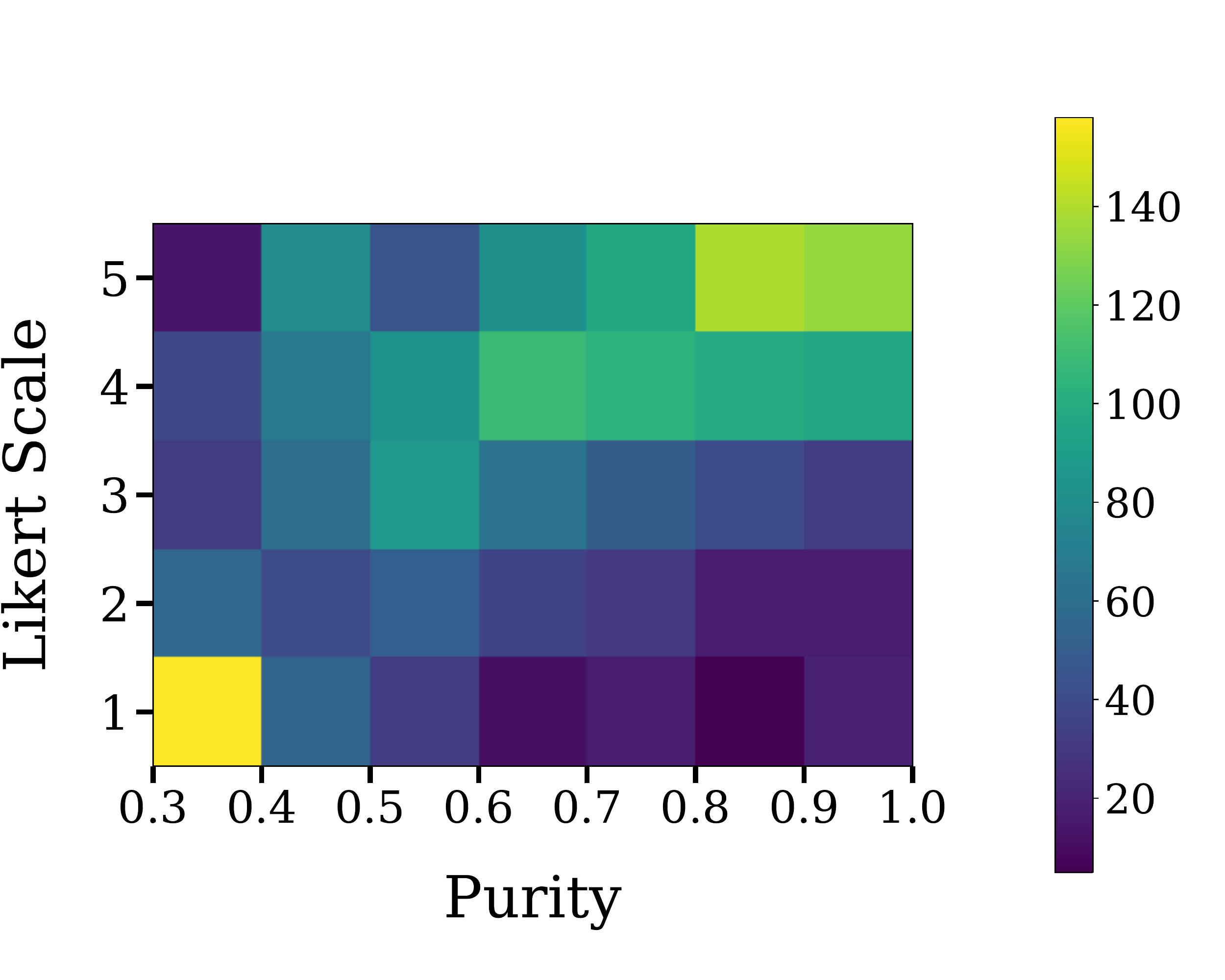}}
\end{subcaptionbox}%
\begin{subcaptionbox}{(b)}[.48\textwidth]
  {\includegraphics[width=.9\linewidth]{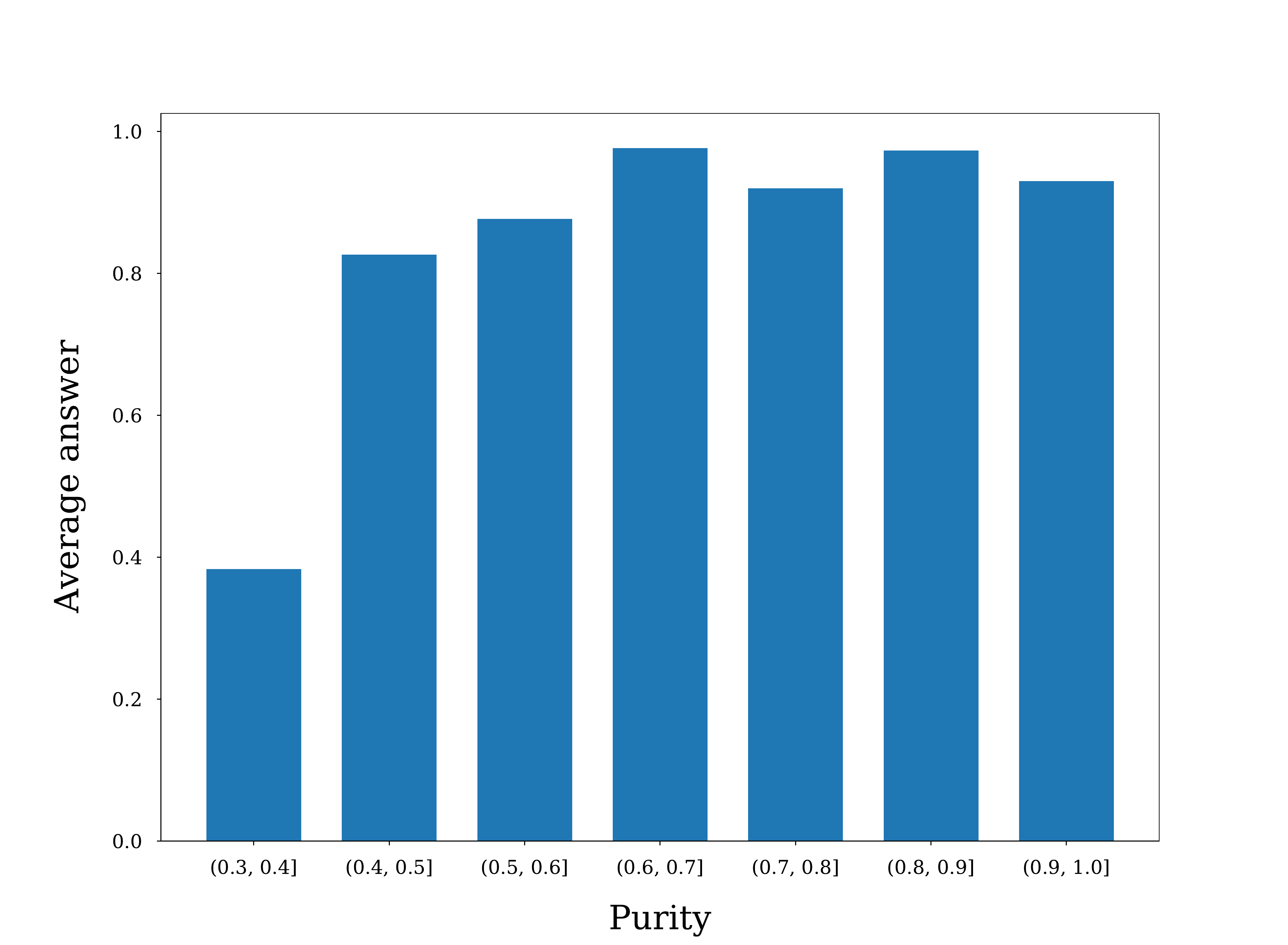}}
\end{subcaptionbox}
\caption{\textbf{Evaluation of class-level caption quality. (a)} 2D histogram of human ratings for the quality of captions for a group as a whole, in a scale from 1 to 5. Each bin summarises all clusters over a purity range, e.g. 0.5--0.6. \textbf{(b)} We also ask whether the provided caption adequately describes at least one image in the group. The plot shows the average answer for clusters in each purity group, (1/yes, 0/no).}
\label{fig:quality}
\end{figure}

In addition to the describability experiments, described in the main paper, we also directly assess the quality of the automatic class-level captions by asking human participants to provide a rating.
Specifically, given a set of images from a class\footnote{The clusters and image sets are the same as those used for measuring learnability (semantic coherence).} and the corresponding class caption, we ask workers to rate the effectiveness of the caption in describing the group \emph{as a whole} using a Likert scale from 1 to 5 (1-Extremely bad, 2-Bad, 3-Adequate, 4-Good, 5-Excellent). 
We also ask whether the description is suitable for \emph{at least one} image in the group, \ie whether it is a partial description (answer:~yes/no), which is particularly meaningful for impure classes. 
The results are shown in \Cref{fig:quality}. We observe that the outcome follows a trend similar to the one shown for describability in the main paper. 
We found that in some cases, class descriptions were rated low, even for highly coherent clusters, due to the inability of the captioning model to identify fine-grained categories in individual images.

\begin{figure}
    \centering
    \includegraphics[width=\textwidth]{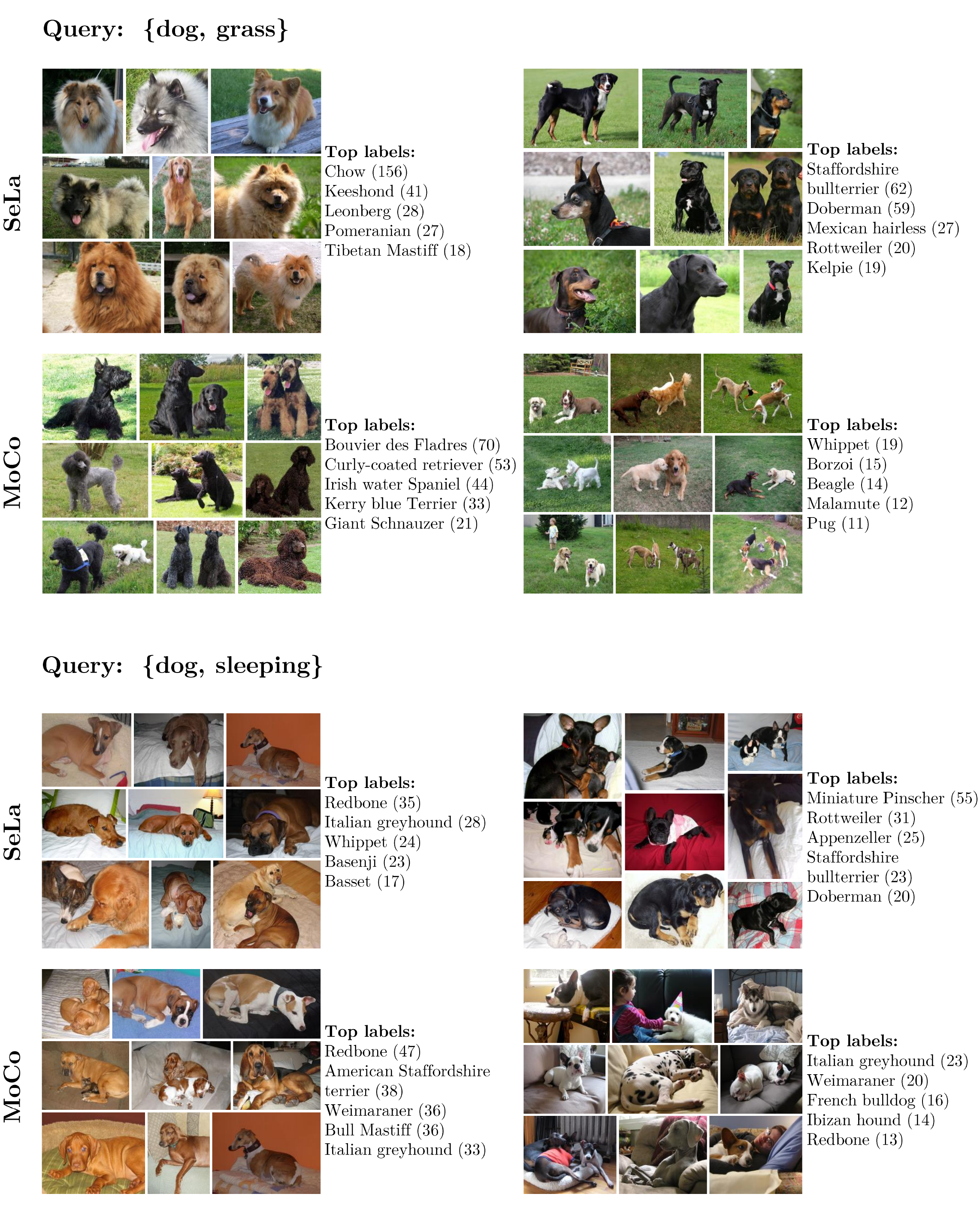}
    \caption{Examples of ``dog'' clusters found by SeLa and MoCo. We observe that \emph{both} methods have learned to group images by color, fur, environment (\eg grass, bed), etc. instead of fine-grained dog breeds.}
    \label{fig:dogs}
\end{figure}

\subsection{Uniqueness of Captions}

To assess the discriminative ability of the captions, we also report the number of unique captions over all 3000 classes, which is 2136 for SeLa and 2070 for MoCo. 
Ignoring stopwords, the number of unique captions becomes 1931 and 1859 respectively.
We report the 10 most frequent class descriptions in \cref{t:freq_cap}.
We then easily observe that these captions correspond to clusters consisting of fine-grained categories, such as breeds of dogs, birds or insects. 
In fact, while some of these clusters are relatively pure, most of them are not, suggesting that self-supervised algorithms in question cannot always learn such fine-grained distinctions.  
As an aside, conventional captioning models cannot exhaustively recognize fine-grained categories either.  
As an example, we show dog clusters in \cref{fig:dogs}, found by searching the class-level descriptions with queries ``dog AND grass" and ``dog AND sleeping".
We verify that the self-supervised algorithms tend to group images by fur, color, environment, activity (\eg playing with other dogs, sleeping, etc.) and even pose or viewpoint rather than distinguishing among dog breeds.  
We again observe similar behavior by both algorithms.

\begin{table}[t]
\centering
\footnotesize
\caption{Most common captions occurring in self-supervised classes.}
\vspace{0.5em}
\small
\begin{tabular}[t]{l p{4.5cm} c @{\hskip 1cm} p{4.5cm} c}
\toprule
Rank & SeLa captions & Count & MoCo captions & Count\\
\midrule
1 & biological species perched on a branch & 27 & biological species perched on a branch & 36 \\
2 & dogs playing in the grass & 21 & dogs playing in the grass & 18 \\
3 & dogs sitting on the floor & 19 & dogs in the grass & 17 \\
4 & a dog in a dog & 18 & a dog with a dog & 16\\
5 & a dog with a dog & 17 & biological species in the grass & 14\\
6 & biological species in the grass & 14 & a dog in a dog & 13 \\
7 & dogs in a field & 11 & dogs sitting in the grass & 13 \\
8 & dogs in the grass & 11 & a monkey sitting in a tree & 12 \\
9 & a grasshopper on a leaf & 11 & a snake on the road & 11  \\
10 & image may contain person playing a musical instrument on stage and indoor & 11 & a spider on a leaf & 11 \\ 
\bottomrule
\end{tabular}
\label{t:freq_cap}
\end{table}

\newpage
\section{Evaluation Details}
In the following, we provide the full list of classes used to collect human judgements on AMT, to quantify learnability and describability.  

\paragraph{Selected ImageNet categories.}
(387) lesser panda,
(145) king penguin,
(685) odometer,
(321) admiral, 
(991) coral fungus, 
(916) website, 
(549) envelope, 
(76) tarantula, 
(807) solar dish, 
(103) platypus, 
(813) spatula,
(731) plunger,
(749) quill,
(910) wooden spoon,
(747) punching bag,
(466) bullet train,
(974) geyser,
(640) manhole cover,
(340) zebra,
(323) monarch.

\paragraph{Selected SeLa classes (sorted by descending purity).} 
2777, 1296, 1537, 933, 1068, 1987, 527, 2434, 396, 2813, 1961, 977, 1332, 1139, 1802, 915, 115, 288, 2047, 136, 184, 1144, 85, 1476, 2375, 1761, 19, 222, 13, 624, 214, 813, 2296, 1993, 1278, 1042, 1406, 1285, 162, 2194, 2090, 1055, 420, 2857, 0, 379, 332, 500, 2250, 39, 63, 1026, 161, 2381, 8, 11, 76, 2233, 2523, 1246, 240, 2258, 338, 2867, 991, 796, 407, 1926, 327, 186. 	

\paragraph{Selected MoCo classes [3k] (sorted by descending purity).}
1386, 252, 844, 1175, 785, 2823, 618, 658, 1501, 1597, 994, 76, 2109, 2960, 240, 924, 1451, 239, 442, 1799, 2881, 1725, 2030, 892, 1243, 370, 2416, 2310, 2102, 2673, 515, 1341, 2595, 2888, 158, 1165, 171, 2964, 608, 238, 350, 1470, 1026, 2612, 2497, 2201, 2992, 253, 169, 2311, 967, 66, 2074, 2003, 2335, 1865, 1730, 2195, 1510, 283, 311, 2490, 1221, 2128, 1206, 1711, 429, 1507, 1601, 410.

\end{document}